\documentclass[conference]{IEEEtran}
\IEEEoverridecommandlockouts

\usepackage{cite}
\usepackage{amsmath,amssymb,amsfonts}
\usepackage{algorithmic}
\usepackage{graphicx}
\usepackage{multirow}
\usepackage{booktabs} 

\usepackage{textcomp}
\usepackage{xcolor}
\def\BibTeX{{\rm B\kern-.05em{\sc i\kern-.025em b}\kern-.08em
    T\kern-.1667em\lower.7ex\hbox{E}\kern-.125emX}}

\begin{document}

\title{Synthetic Data Generation and Deep Learning for the Topological Analysis of 3D Data\\
\thanks{This research was supported by the Australian Government through the ARC's Discovery Projects funding scheme (project DP210103304). The first author was supported by a Research Training Program (RTP) Scholarship – Fee Offset by the Commonwealth Government.

Copyright 2023 IEEE. Published in the Digital Image Computing: Techniques and Applications, 2023 (DICTA 2023), 28 November – 1 December 2023 in Port Macquarie, NSW, Australia. Personal use of this material is permitted. However, permission to reprint/republish this material for advertising or promotional purposes or for creating new collective works for resale or redistribution to servers or lists, or to reuse any copyrighted component of this work in other works, must be obtained from the IEEE. Contact: Manager, Copyrights and Permissions / IEEE Service Center / 445 Hoes Lane / P.O. Box 1331 / Piscataway, NJ 08855-1331, USA. Telephone: + Intl. 908-562-3966.}
}

\author{\IEEEauthorblockN{Dylan Peek}
\IEEEauthorblockA{\textit{School of Information and Physical Sciences} \\
\textit{The University of Newcastle}\\
Callaghan NSW 2308, Australia \\
Dylan.Peek@uon.edu.au\\
OrcidID: 0000-0003-3568-7853}
\and
\IEEEauthorblockN{Matthew P. Skerritt}
\IEEEauthorblockA{\textit{Mathematical Sciences} \\
\textit{RMIT University}\\
Melbourne VIC 3000, Australia \\
Matt.Skerritt@rmit.edu.au\\
OrcidID: 0000-0003-2211-7616}
\and
\IEEEauthorblockN{Stephan Chalup}
\IEEEauthorblockA{\textit{School of Information and Physical Sciences} \\
\textit{The University of Newcastle}\\
Callaghan NSW 2308, Australia \\
Stephan.Chalup@newcastle.edu.au\\
OrcidID: 0000-0002-7886-3653}
}

\maketitle

\begin{abstract}
This research uses deep learning to estimate the topology of manifolds represented by sparse, unordered point cloud scenes in 3D. 
A new labelled dataset was synthesised to train neural networks and evaluate their ability to estimate the genus of these manifolds. This data used random homeomorphic deformations to provoke the learning of visual topological features.
We demonstrate that deep learning models could extract these features and discuss some advantages over existing topological data analysis tools that are based on persistent homology. Semantic segmentation was used to provide additional geometric information in conjunction with topological labels. Common point cloud multi-layer perceptron and transformer networks were both used to compare the viability of these methods.
The experimental results of this pilot study support the hypothesis that, with the aid of sophisticated synthetic data generation, neural networks can perform segmentation-based topological data analysis. While our study focused on simulated data, the accuracy achieved suggests a potential for future applications using real data.
\end{abstract}
\section{Introduction}
Topology is the mathematical field that studies topological shape which is invariant to scale and homeomorphic deformations~\cite{Hatcher2002}. In its applied form Topological Data Analysis (TDA) can offer insight into topological features of data such as the amount of voids or n-dimensional holes~\cite{EdelsbrunnerHarer2010,otter2017roadmap}. Recent theoretical and computational advancements have allowed a rapid uptake in real-world use. It has been successfully used in Chemical Engineering~\cite{smith2021topological}, EEG Processing~\cite{xu2021topological}, Medical Diagnosis~\cite{yamanashi2021topological,rucco2014using}, Aviation~\cite{li2019topological}, Nuclear Physics~\cite{hamilton2022applications}, Heart Rate Variability Analysis~\cite{chung2021persistent} and Network Analysis~\cite{carlsson2021topological}.

Persistent homology is a powerful tool for TDA that can be used to produce persistence intervals of topological features across different dimensional scales~\cite{otter2017roadmap,bubenik2007statistical}. These intervals can be expressed in forms such as persistence diagrams, persistence images and persistence barcodes~\cite{adams2017persistence}. Analysis of these intervals can offer insight into the topological features of data, this data may be expressed in n-dimensional volumetric, mesh or point cloud structures. 

We propose that neural networks can be used as an alternate strategy to some of the topological aspects of persistent homology where it may offer advantages that have yet to be leveraged in research spaces.

One limitation when extracting topological signatures such as Betti numbers via interval analysis, as in persistent homology, is that the conversion of raw data into intervals may be a lossy process~\cite{Hatcher2002,otter2017roadmap,EdelsbrunnerHarer2010}. 
An extraction of n-dimensional holes and voids from intervals can induce some ambiguity in the original configuration. 

\begin{figure}[htbp]
    \centering
    \begin{tabular}{|c|c|}
        \hline
        \raisebox{-1.1cm}{\includegraphics[trim=0 300mm 0 250mm, clip, width=0.2\textwidth]{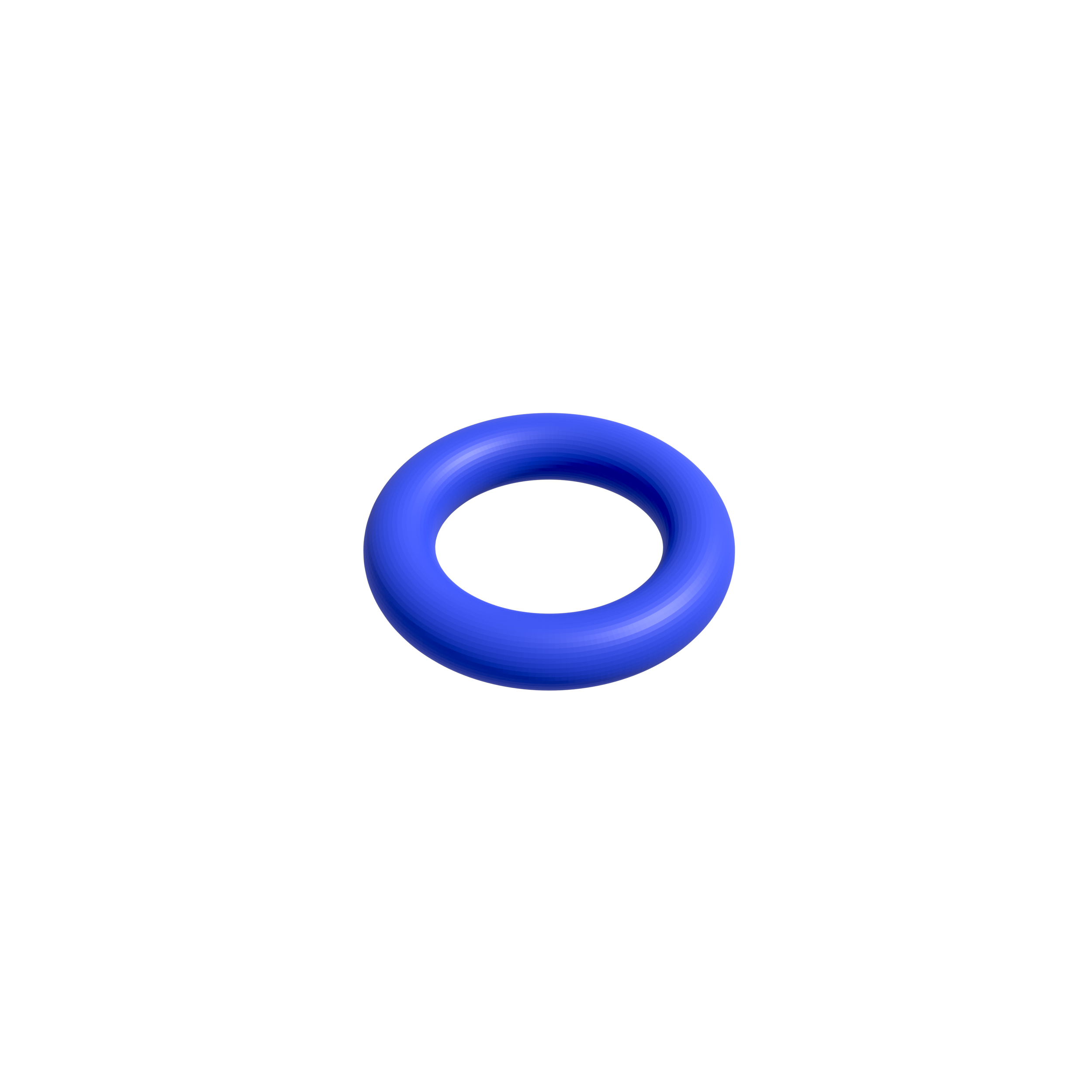}} &
        \raisebox{-1.1cm}{\includegraphics[trim=0 300mm 0 250mm, clip, width=0.2\textwidth]{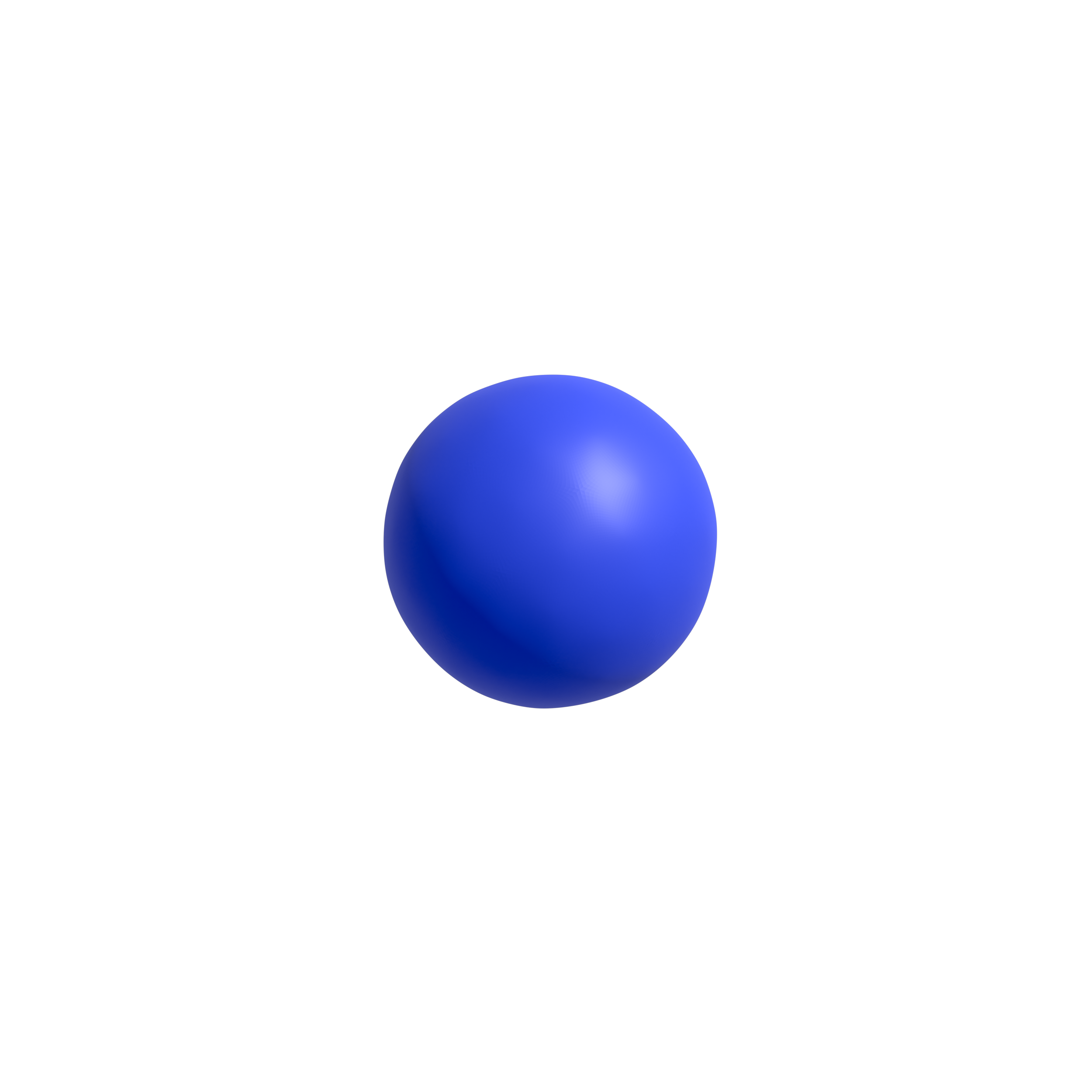}} \\
        \includegraphics[trim=0 250mm 0 300mm, clip, width=0.2\textwidth]{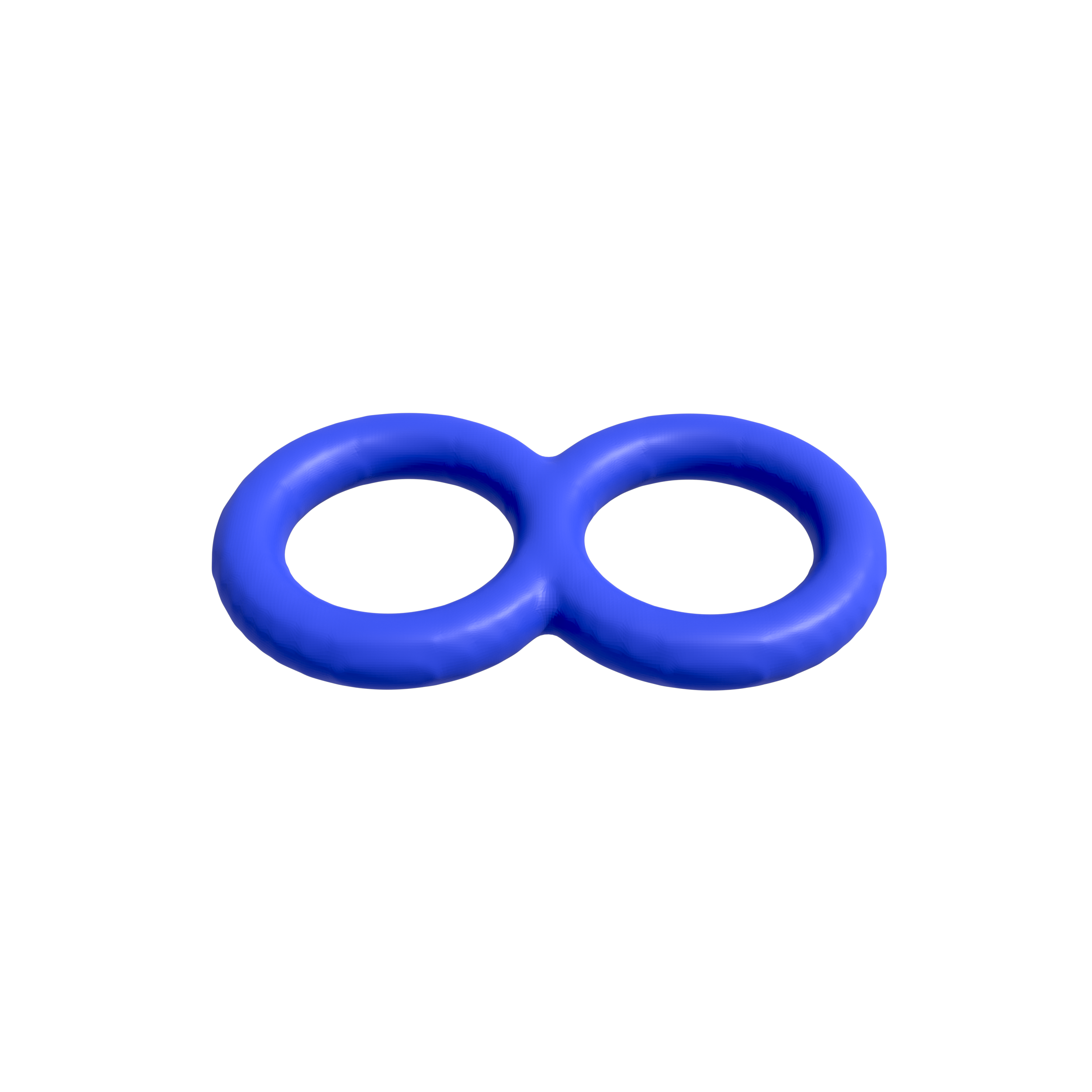} &
        \includegraphics[trim=0 250mm 0 300mm, clip, width=0.2\textwidth]{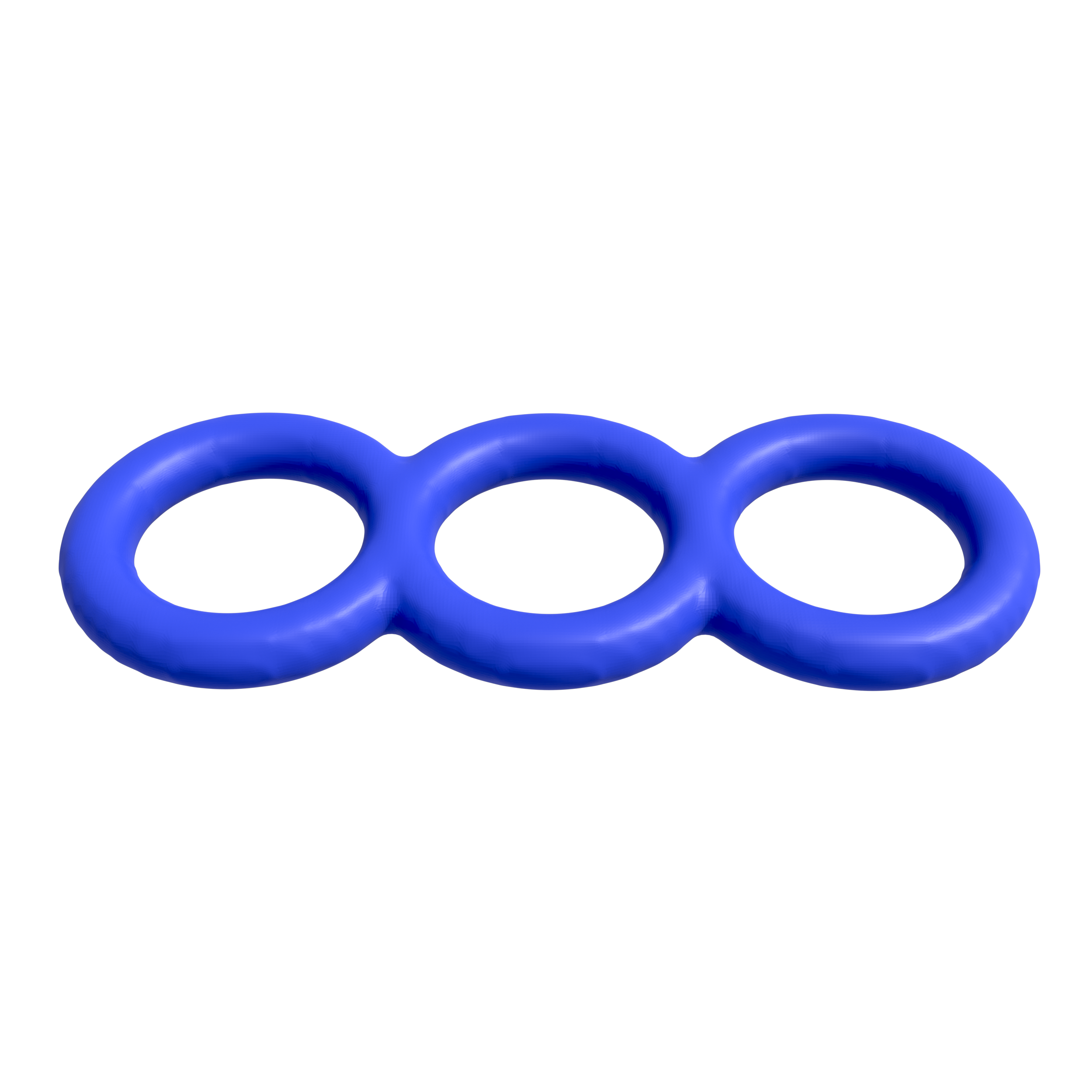} \\
                \hline
        \addlinespace[0.25cm]
    \end{tabular}
\caption{Left and right are two scenes demonstrating potential ambiguities when assessing the global topology of a scene using persistent homology. Both scenes have the Betti numbers \(\beta_0 = 2\), \(\beta_1 = 6\), and \(\beta_2 = 2\).
 The genus and Betti numbers for these objects can be seen in Table \ref{background:topology:table}. \\The meshes of the four displayed surfaces are examples of genus $\left\{0,1,2,3\right\}$ manifolds that were used as seeds in the data generation process, see Section~\ref{dataset}. The visualisations were performed in Blender~\cite{blender3.0.1}.}
\label{background:topology:bettifig}
\end{figure}

For example, a scene may comprise 2 objects and 3 holes however it is ambiguous whether the scene contained a 0-hole object and a 3-hole object or a 1-hole object and a 2-hole object, see Fig~\ref{background:topology:bettifig}. 
Our approach is based on per-point labelling segmentation which addresses such ambiguities by attempting to isolate these objects into classes. 

There are other potential advantages that neural networks could provide to researchers. Persistent homology can have large memory and CPU requirements and a time complexity that scales with point cloud sample counts~\cite{malott2022survey}. These computational constraints may prevent utilization in certain applications where the resources or time are not available. Details regarding some key concepts and algorithms for the computation of persistent homology are surveyed in~\cite{otter2017roadmap} which compares time, CPU, and memory benchmarks across these methods. 
Neural networks may also offer an intelligent interpretation of noise, density, and connectivity. 
Persistent homology can be robust to noise that has known characteristics however may incur problems when evaluating data with varied noise, relative object scale, and sample density~\cite{turkevs2021noise,otter2017roadmap}. Neural networks may use the local and global contexts of points and surfaces to intelligently extract topological features based on `human-like' intuition.

In this paper we analyse closed compact orientable 2-dimensional manifolds to estimate the topological `genus'.
Each of these manifolds is homeomorphic to a sphere with a number of handles attached, where the genus $g$ is the number of these handles~\cite{gallier2013guide}. The genus is related to the Euler characteristic $\chi$ and the first Betti number $\beta_1$ by the formula: $g = \frac{1}{2} \beta_1= \frac{3}{2} \chi$~\cite{EdelsbrunnerHarer2010}. 
Table~\ref{background:topology:table} lists some example surfaces including their genus, Betti numbers, and Euler characteristic.

We propose employing semantic segmentation neural networks to directly extract genus information from raw data, providing a novel tool for TDA. To our knowledge, no research exists to evaluate neural networks’ ability to extract Betti numbers and/or genus from raw 3D point cloud data, nor to apply segmentation labelling for topological signatures. Networks aimed at topological data analysis with different inputs, outputs, and objectives are discussed in Section~\ref{relatedworks}.

An investigation into existing datasets and related literature found no suitable labeled sets for our application of topological machine learning, some of these datasets are discussed in Section~\ref{relatedworks}. 
Our dataset required Betti number or genus labels, scenes that included multiple objects and genus groups, and variability between these objects appearance. Real-world data is valuable for certain applications however we aimed to isolate the `human-like' element of visual topological analysis. For this, varied homeomorphic deformation for objects with the same topological signatures was deemed necessary to incentivise the learning of target features.

\begin{figure}[htbp]
\centering
\begin{tabular}{@{}c@{\hspace{.5em}}c@{}}
  \includegraphics[height=40mm,width=35mm]{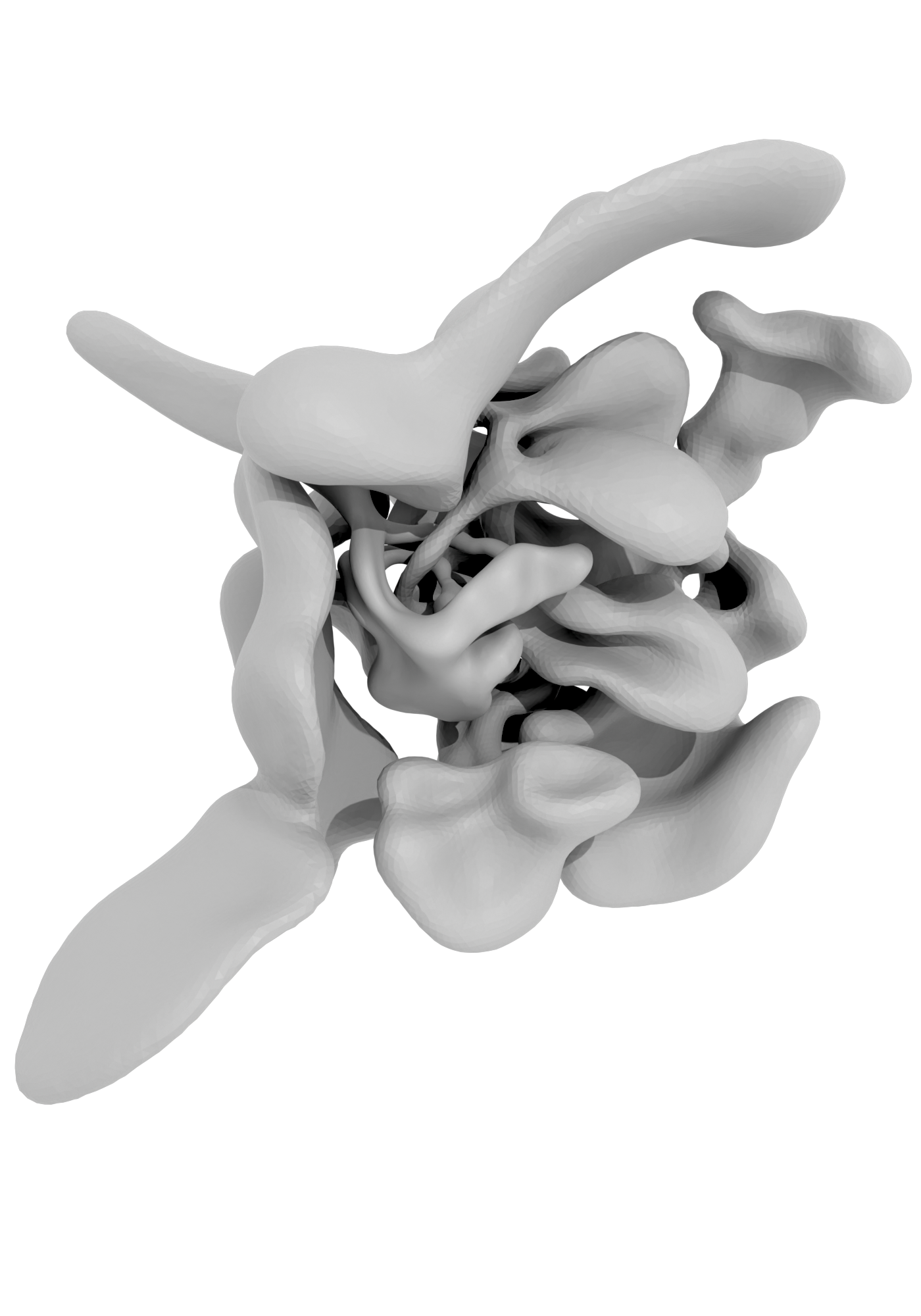} &
  \includegraphics[height=40mm,width=35mm]{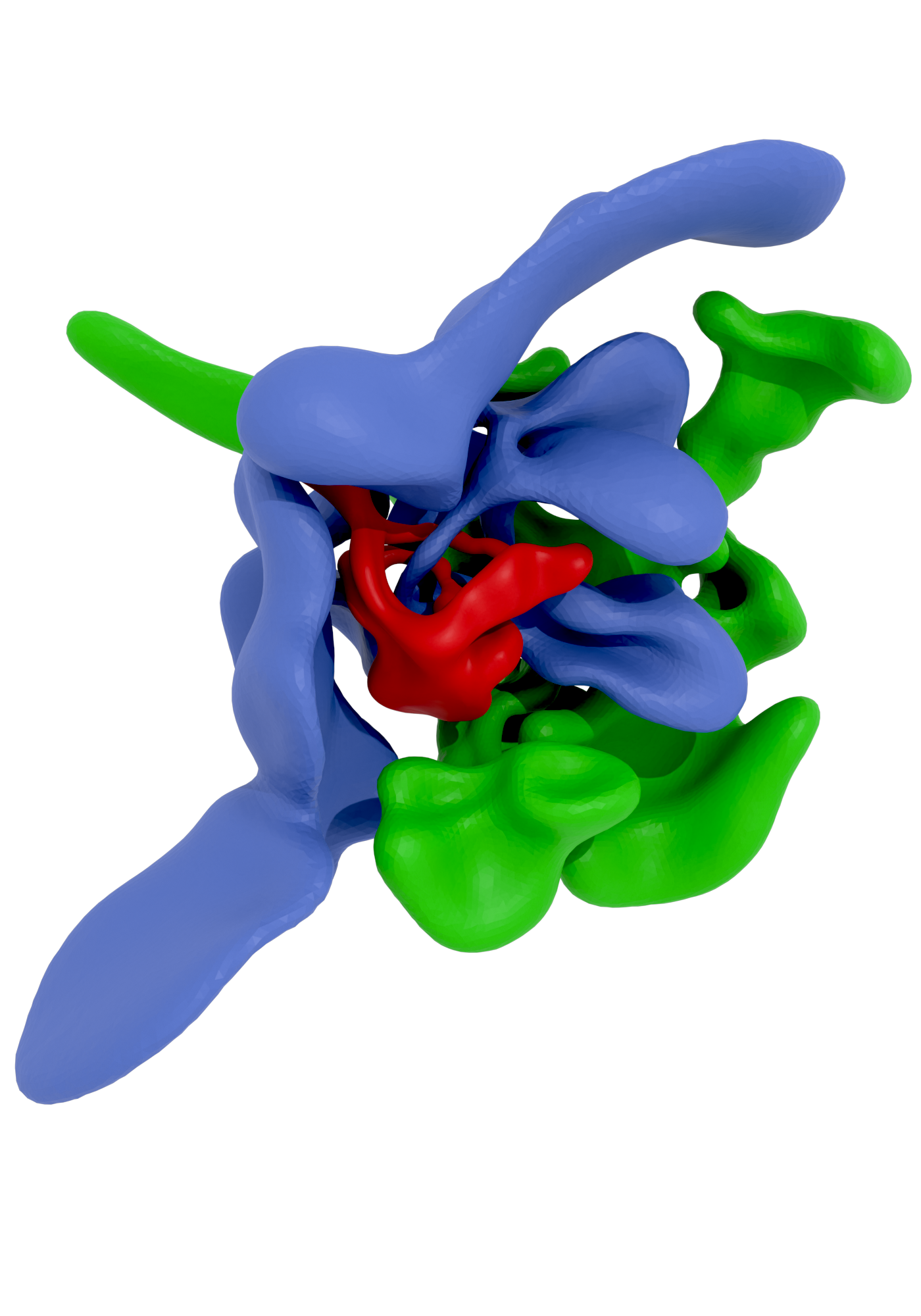} \\
  (a) & (b)
\end{tabular}
\caption{Sample comprising of 3 close proximity objects of known genus. Image (a) consists of a generated dataset mesh. Image (b) has been coloured strictly for demonstrative purposes; genus 0 (blue), genus 2 (green) and genus 3 (red). Visualisation of data was performed in Blender 3.0.1. \cite{blender3.0.1}}
\label{intro:datasetexample}
\end{figure}

To fulfill our criteria we created a new dataset using a combination of the Repulsive Surfaces algorithm~\cite{yu2021repulsive} and the Wave Function Collapse algorithm~\cite{gumin2016wavefunctioncollapse} to produce 3D meshes of known properties. We refer to this dataset as the `Repulse' dataset. An example of an element in the dataset is shown in Fig~\ref{intro:datasetexample} which demonstrates how visually challenging it may be for humans to topologically analyze data that poses little correlation to our existing experiences.

\begin{figure*}[htbp]
\centerline{
{\includegraphics[height=50mm,width=40mm]{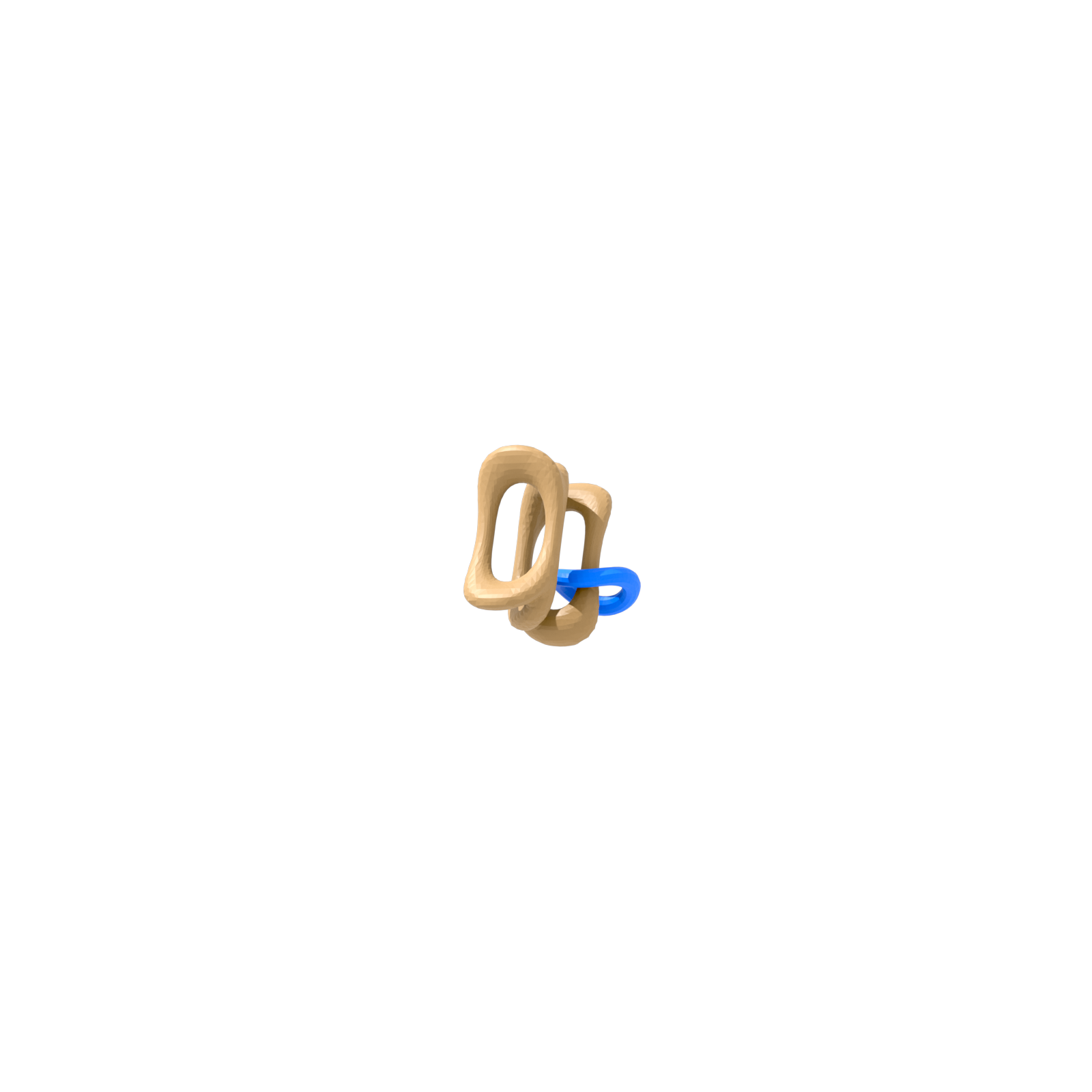}}
{\includegraphics[height=50mm,width=40mm]{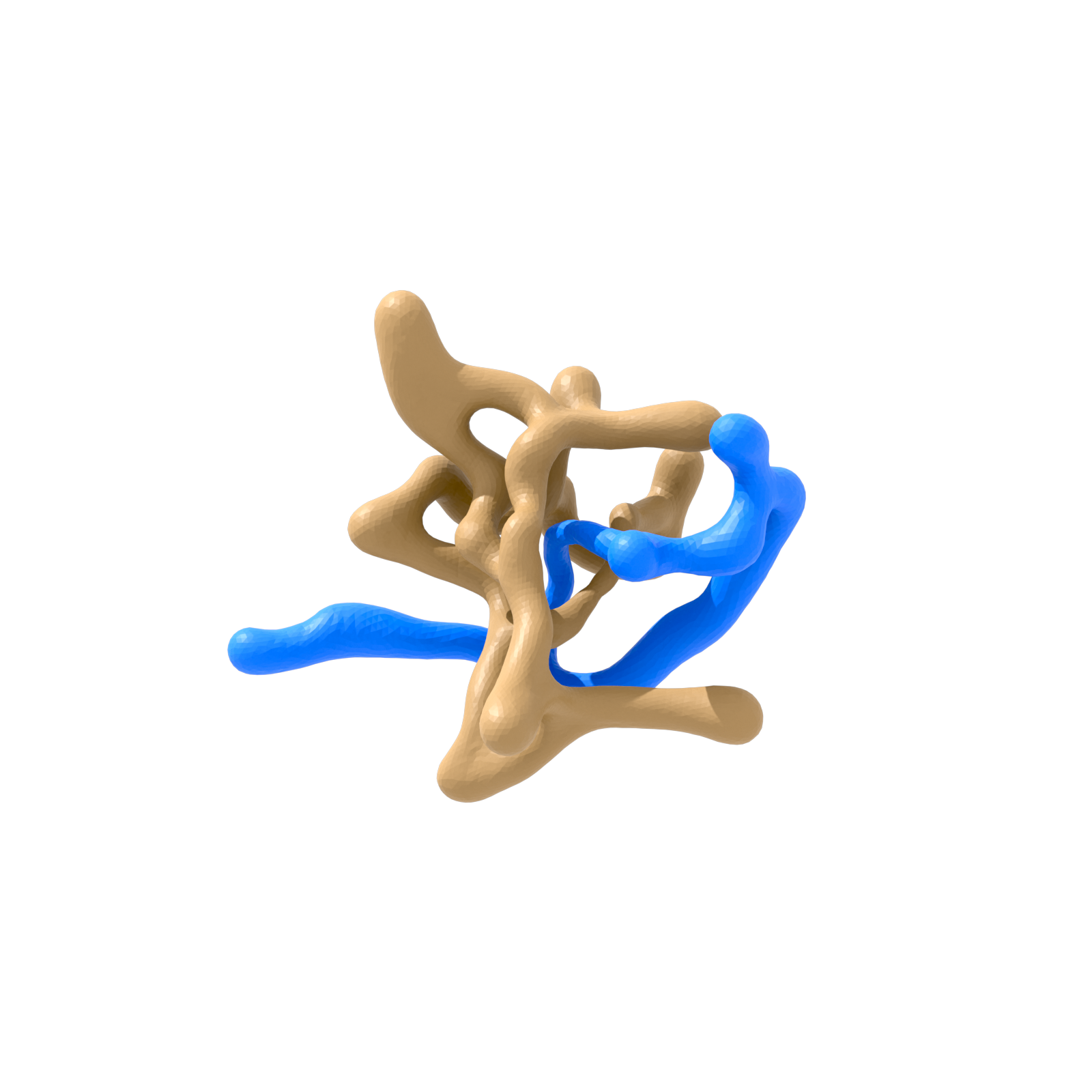}}
{\includegraphics[height=50mm,width=40mm]{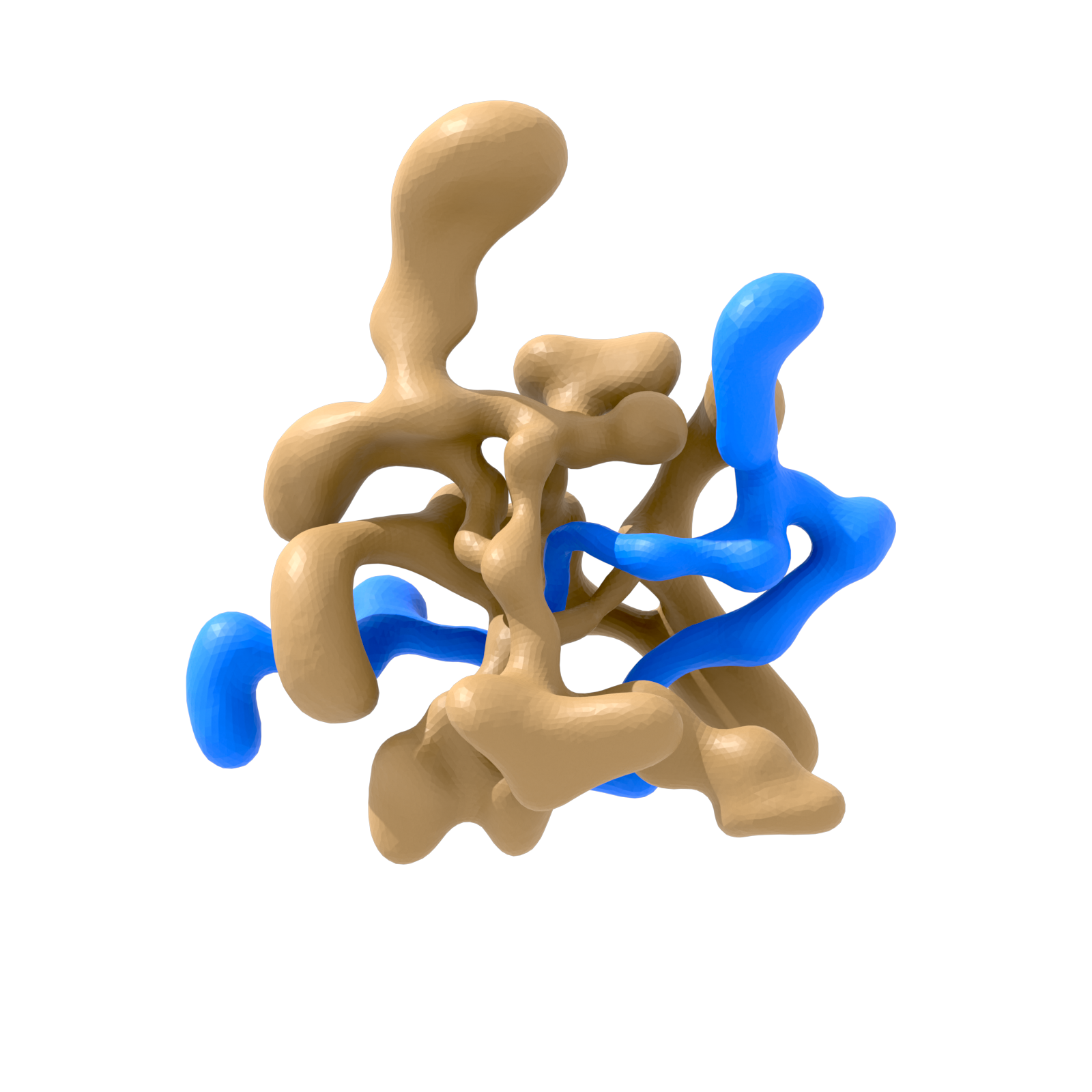}}
{\includegraphics[height=50mm,width=40mm]{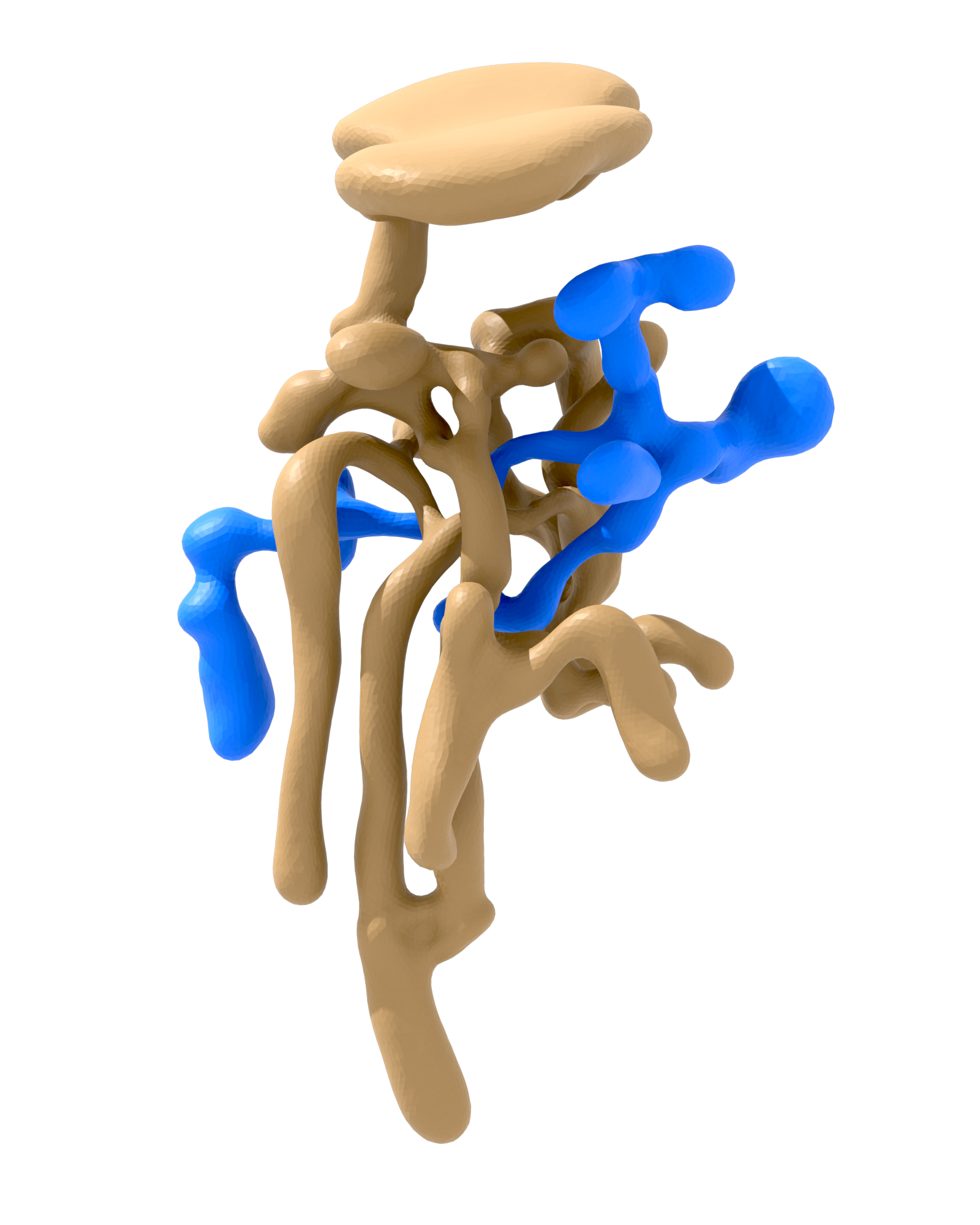}}}
\caption{Sample showing the growth of 2 interlinked objects of genus 1 (blue) and genus 3 (brown). Visualisation of data was performed in Blender 3.0.1.~\cite{blender3.0.1}}
\label{fig:datasetgrowth}
\end{figure*}

The key contributions of this research are:
\begin{enumerate}
  \item Generation of a novel labeled dataset with scenes of varying object count and genus that could be used for topological machine learning training and evaluation. This dataset can be used in its original mesh form or sampled in point clouds of various point counts.
  \item Training multiple neural network models to analyze 3D point cloud data and assign each point a genus corresponding to its estimated object grouping.
  \item Discuss the viability of novel topological data semantic segmentation and compare network performance with respect to architectural design.
\end{enumerate}

The dataset generation process is described in Section \ref{dataset} and a brief visual summary can be seen in Fig \ref{fig:datasetgrowth}.

\subsection{Related Works} \label{relatedworks}
Recent studies~\cite{paul2019estimating, HannouchChalup2023} have shown success in Betti number estimation of 2D, 3D, and 4D volumetric data. They used synthetic data comprising a range of fundamental topological shapes without deformation in pixel, voxel, and toxel form and conducted TDA with basic convolutional neural networks. 

In contrast, the present project focuses on increasing the complexity of data in 3D significantly by generating sophisticated datasets with diverse organic structures. 
By switching from a classification to a segmentation paradigm and using more advanced deep learning models we can demonstrate that more information can be drawn from the resulting network output.
The long-term vision of this project is to achieve better manageable computational performance and real-world applicability of this approach.

Two networks have emerged to produce estimated persistence images of various input data formats: Pi-Net, and TopologyNet. Persistence images are a representation of persistent diagrams (see Section~\ref{ssub:background:toplogy}) in $\mathbb{R}^n$ space. Further mathematical and computational detail on the creation of persistence images can be seen in~\cite{adams2017persistence}.

Pi-Net~\cite{som2020pi} introduced two convolutional neural networks to produce estimated persistence images of multivariate and multichannel input data. Signal PI-Net can process time series signals and Image PI-Net can process 2D image data. 
An output format of persistence images still requires further subjective interpretation for the extraction of additional topological features. In this form, however, raw data information that may be useful in determining these features is lost. A one-hit network for outputting a variety of dimensional features could provide additional information. For these 2D image data experiments the datasets CIFAR10~\cite{krizhevsky2009learning}, CIFAR100~\cite{krizhevsky2009learning} and SVHN~\cite{netzer2011reading} were used. These datasets use a combination of common objects and street-view house numbers. As these datasets were not exposed to homeomorphic deformation there may exist a correlation between the geometric shape classification and topological labels.

A TopologyNet model was used to fit 3D point cloud data directly to persistence images~\cite{zhou2022learning}. The training of this network used ModelNet40 dataset~\cite{wu20153d} comprising 40 categories for classification such as a chair or table. These classes may have spurious correlations between object type and topological labels making them undesirable for evaluating topological models.  

RipsNet~\cite{de2022ripsnet} model, like TopologyNet, also successfully used 3D point cloud data however outputs the persistence diagram instead of a persistence image. The creation of persistence diagrams was identified as the most computationally expensive process in the extraction of features and thus estimating this with neural networks can greatly reduce computational requirements. Ripsnet training used ModelNet10 dataset~\cite{wu20153d} comprising 10 object categories. This shares the topological limitations identified for ModelNet40. 

Further testing on the topological space evaluation abilities of neural networks is performed in~\cite{montufar2020can}. 
This paper emphasizes the benefit of directly outputting topological features rather than persistence diagrams in certain applications to bypass additional computational processing. This work was successful in estimating binary features and ``tropical coordinates'' from 2D images.
In many applications, the persistence diagram is nothing more than a necessary intermediary step in the calculation of topological features such as Betti numbers. 

Our research also attempts to bypass persistence diagram creation and instead directly outputs topological features as labels for each point in the cloud. This allows for object extraction filtered by genus or Betti numbers to produce shortlisted key data subsets. 

The topological property on which we segment the network is genus (or, equivalently, Betti number) which we discuss in Section~\ref{ssub:background:toplogy}.

Our motivation for this research extends beyond achieving high accuracy in topological analysis. We aim to explore the inherent capacity of neural networks to analyse topological signatures without leveraging pre-learned assumptions regarding object class. 
Pre-existing literature uses machine learning for the interpretation of persistence intervals and the creation of persistence diagrams/images. Much of this work uses datasets with spurious correlations between geometric structure and topological features. Prior knowledge regarding certain object characteristics can make it challenging to isolate the true ability of neural networks to interpret data from a topological perspective. We attempt to mitigate this issue through random homeomorphic deformations when generating the Repulse dataset. 

We mention in passing that leveraging domain-specific correlations to enhance accuracy may be desirable in real-world applications. However, for our study it compromises the conceptual purity and generality of our approach.

Furthermore, there is an adjacent line of research where the lifespan of persistence intervals can be analyzed to extract topological signatures such as Betti numbers. This interval analysis can be subjective and case specific as different samples and applications may have different properties and scales. Existing research has successfully shown that neural networks can be beneficial in interpreting persistence intervals~\cite{hajij2021tda,hofer2017deep,rostami2019deep}. These networks utilize persistent homology as a pre-processing stage and are subsequently still prone to the resource, time, and ambiguity factors previously discussed. 

\subsection{Approach, Data, and Networks}
With no existing research, experiments, or benchmarks for 3D topological segmentation, it was deemed premature to introduce additional architectural changes or propose a new network without pre-existing benchmarks. 
Instead, we apply a novel approach to TDA via feature segmentation and introduce a synthetic dataset designed to train and evaluate this approach with selected existing networks.

The survey~\cite{guo2020deep} explores the leading networks for deep learning on 3D point cloud data.

PointNet~\cite{charles_su_kaichun_guibas_2017} was pioneering research that introduced a multi-layer perceptron network for 3D point cloud data classification and segmentation. PointNet performs well at extracting global features and works on each point individually making it input permutation invariant. Due to this invariance to point order, PointNet does not perform competitively at extracting features at local scales.

To increase local feature extraction a hierarchical network PointNet++ was proposed by~\cite{qi2017pointnet++}.
PointNet++ expands upon PointNet and uses a 3 abstraction layer approach to analyze multiple feature scales.

Multiple networks have emerged to further improve the accuracy of 3D point cloud deep learning. As there are not sufficient benchmarks for topological analysis we considered top performing and intriguing networks evaluated on the S3DIS dataset~\cite{armeni20163d}. This dataset contains large, indoor office space data with labelling for structural elements and common furniture and items. Comparing network accuracy on the S3DIS dataset would not directly translate to this new topological task, however, it gave some indication of relative performance for 3D segmentation.  

The three networks chosen for our experiments were PointNet++, RepSurf-U~\cite{ran2022surface}, and Point Transformer~\cite{zhao2021point}.

The Umbrella RepSurf (RepSurf-U) network~\cite{ran2022surface} is built upon PointNet++ and is inspired by umbrella curvature~\cite{foorginejad2014umbrella} concepts from computer graphics to form explicit local connections in raw unordered point cloud data. This curvature construction can offer surface and connectivity context to otherwise sparse, edgeless point sets.

The Point Transformer network~\cite{zhao2021point} introduces point transformer blocks with self-attention layers for 3D point cloud processing. Self-attention~\cite{vaswani2017attention} attempts to assess the contextual weight of embeddings such as the significance of words in a sentence with respect to a target word. Point Transformer applies self-attention around centroid datapoints within point subsets known as `local neighbourhoods' which are formed via k-nearest neighbours.

The accuracy of these networks when evaluated on S3DIS dataset (Area-5) can be seen in Table~\ref{introduction:networkacc:table}. These results were obtained from the official RepSurf GitHub repository~\cite{ran2022repsurf}.
\begin{table}[h]
\caption{Accuracy of chosen networks performed on S3DIS (Area-5).} \label{introduction:networkacc:table}
\centering
\begin{tabular}{lccc}
\toprule
Model & mIoU & mAcc & OA \\
\midrule
PointNet++ & 64.05 & 71.52 & 87.92 \\
RepSurf-U & 68.86 & 76.54 & 90.22 \\
Point Transformer & 70.4 & 76.5 & 90.8 \\
\bottomrule
\end{tabular}
\end{table}

PointNet++ was used in our experiments as it is the backbone and inspiration for many new architectures which served as an accuracy baseline. While it scored lower than RepSurf-U and Point Transformer, it was necessary to evaluate whether the enhancements in newer networks were equally applicable to TDA or if they were specifically optimized for recognition segmentation tasks.
RepSurf-U was chosen as it uses PointNet++ as a backbone with specific design elements for surface construction which seemed conceptually interesting for manifold analysis.
Point Transformer was chosen as there is emerging research interest around transformer networks, especially in the natural language processing field. We found it compelling to assess how such architectures fare in TDA tasks. 

\section{Background}
\subsection{Three-Dimensional Point Cloud}
Three-dimensional point cloud data comprises of points in space with $x$, $y$, and $z$ coordinates and may include other channels such as red, green, and blue color (RGB).
Point clouds contain no information about connectivity between points; only the points themselves are present.

Point cloud data is commonly used as many real-world sensors such as depth imaging and lidar construct scans based on distance to the surface of an object. Additionally, it is possible to sample existing mesh or voxel data into a point cloud with varying point counts.

We note that sometimes network training with point cloud data will also include the normal vector to the surface of the object approximated by the point cloud input. This information was not utilised for experimentation on the Repulse dataset as it offers additional boundary and surface context. For complex scenes with unordered and close proximity points the normal vectors may be hard to determine. Additionally, these vectors are not used in conventional persistent homology.

\subsection{Topology} \label{ssub:background:toplogy}
Geometric Analysis can describe properties in Euclidean space such as size, shape and volume.
Such geometric properties change when objects are deformed within the space.
Topological Analysis, on the other hand, studies properties that are invariant to homeomorphic deformations and instead describes fundamental structural features such as holes and voids.
Homology is a key concept in topological analysis (see e.g.,~\cite{Hatcher2002}).
For a manifold, $X$, and every dimension, $n$, there exists a vector space $H_n(X)$---called the $n$\textsuperscript{th} Homology group of $X$---which characterizes the topological $n$-dimensional components of the space $X$.

A related notion to the homology groups are the Betti Numbers, ${\beta_n}$.
Formally, the $n$\textsuperscript{th} Betti number of a topological space, $X$, is the dimension of the $n$\textsuperscript{th} homology group of $X$.
Informally, the  $n$\textsuperscript{th} Betti number describes the $n$-dimensional ``holes'' within a manifold. In 3-dimensional Euclidean space only the first three Betti numbers are relevant:
\begin{itemize}
    \item ${\beta_0}$ can be conceptualized as the number of disconnected objects or blobs, 
    \item ${\beta_1}$ can be conceptualized as the number of 1-dimensional holes, i.e., those bounded by $S^1$.
    \item ${\beta_2}$ can be conceptualized as the number of voids or bubbles in 3-dimensional space.
\end{itemize}
Through this definition, a coffee mug has the same Betti numbers as a ring, and glasses frames have the same as a figure 8. Examples of Betti numbers for different manifolds can be seen in Fig~\ref{background:topology:bettifig}.

We mention in passing that the Euler characteristic of a manifold is related to the Betti numbers through the equation:
\[ \chi = \beta_0 - \beta_1 + \beta_2 - \beta_3 + ... + \beta_n \]

Our topological Repulse dataset comprises of compact orientable surfaces which consist solely of spheres, $S^2$ and connected sums of $g$ tori, $T^2 \# ... \# T^2$, where $g$ is the genus, a natural number (i.e., $g \in \mathbb{N}$) denoting the number of holes. The Betti numbers and Euler characteristics for $g \in \left\{0,1,2,3\right\}$ are shown in Table \ref{background:topology:table}

\begin{table}[h]
\caption{Genus ($g$), Betti numbers ($\beta_n$) and Euler characteristic ($\chi$) of closed compact orientable surfaces.} \label{background:topology:table}
\centering
\begin{tabular}{lccccc}\toprule 
Surface $M$ &$g$&$\beta_0$&$\beta_1$&$\beta_2$&$\chi$\\[3pt]\midrule
Sphere  $S^2$&0&1&0&1&2\\
Torus $T^2$&1&1&2&1&0\\
2-holed torus $T^2 \# T^2$&2&1&4&1&-2\\
3-holed torus $T^2  \# T^2 \# T^2$&3&1&6&1&-4\\
$g$-holed torus $T^2\# \dots \# T^2$&$g$&1&2$g$&1&2-2$g$\\[3pt]\bottomrule
\end{tabular}\label{table:bettis}
\end{table}

The dataset is (as previously stated) restricted to compact orientable surfaces in 3 dimensions. This eliminates non-orientable surfaces such as M\"{o}bius strips and Klein bottles which have holes but no internal closed volumes. Under these restrictions, the Betti numbers can be calculated from the collective genus values of objects in the scene. This relationship is not bijective and subsequently the inverse can not be calculated as different permutations of objects can produce the same Betti numbers. The example scene mentioned in the introduction comprises a torus and a 2-holed torus and consequently has $\beta_0 = 2$, $\beta_1 = 6 $ and $\beta_2 = 2$ as does a scene with a sphere and a 3-holed torus. This made per-object genus labeling for the Repulse dataset a more favourable descriptor. It should be noted that the ability to deal with this ambiguity associated with the Betti numbers is an advantage of our proposed approach using neural networks over traditional persistent homology methods.

For more detailed information on the algebraic topology and statistical concepts of persistent homology, see~\cite{bubenik2007statistical}. A survey and evaluation of common methods for the computation of persistent homology can be found in \cite{otter2017roadmap}.

\section{Method} \label{method}

\subsection{Dataset} \label{dataset}
As previously stated, our dataset was created using a combination of the Repulsive Surfaces algorithm \cite{yu2021repulsive} and the Wave Function Collapse algorithm~\cite{gumin2016wavefunctioncollapse}.

The Repulsive Surfaces algorithm uses repulsive tangent point energies of mesh data to perform collision avoiding optimizations on 3D objects. This optimization attempts to simplify homeomorphic deformations to represent the object in a topologically equivalent form. The strength of this algorithm for our application in topological data generation is that it provides a barrier to intersections, which prevents the genus from changing throughout the surface deformation. 

Instead of the typical use in reducing homeomorphic deformation we used the algorithm to \emph{induce} deformation by increasing the objects' surface area within random environmental constraints. Some basic seed objects for $g = 0, 1, 2 ,3$ are displayed in Fig~\ref{background:topology:bettifig}.


The wave function collapse algorithm allows for random arrangement of rule-based tiles. This was used to create 3D environments with random shape for the objects to grow in.

\begin{figure}[htbp]
\centerline{{\includegraphics[trim=0 140mm 0 200mm, clip, scale=0.3,height=35mm]{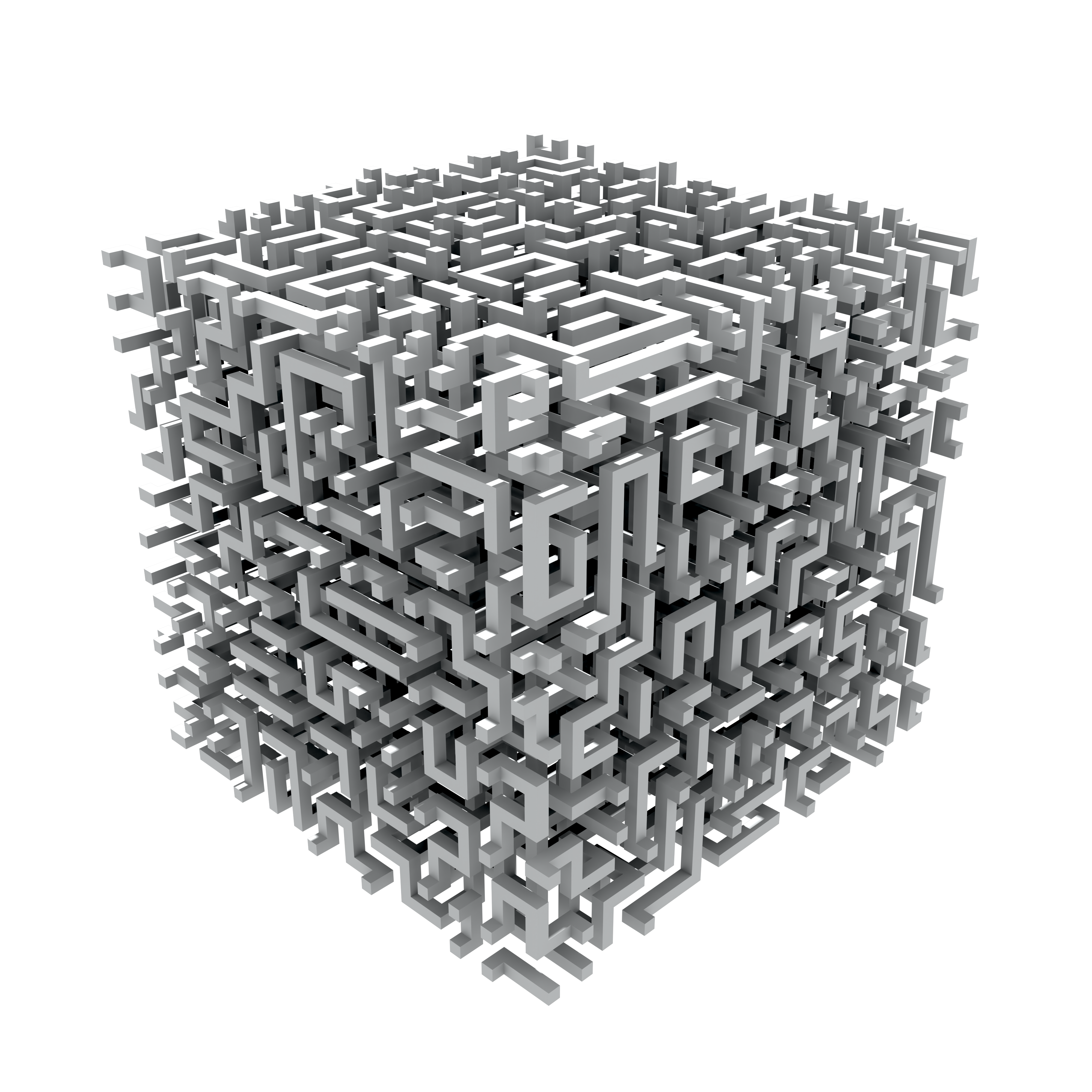}}}
\caption{An environment generated using the wave function collapse algorithm.}
\label{dataset:wfcenv}
\end{figure}

Each 3D tile comprised of 3x3 cells with a boolean value indicating cell presence. The 3x3 cells could form a variety of shapes such as straight lines, 90 degree turns, and intersections that when assembled produced a 3D maze-like structure that acted as a restrictive barrier. An example randomly generated environment is shown in Fig~\ref{dataset:wfcenv}.

Using this method the seed objects could be randomly grown through a series of algorithmic iterations providing unique deformations. See Fig~\ref{fig:datasetgrowth} for an example of dataset growth. This iterative growth process allowed sampling at various stages to produce a dataset with varied complexity. Growing of seed objects within a confined space is conceptually similar to vines or plants growing around a wooden lattice where weaving and wrapping creates unique shapes. 

\begin{figure}[htbp]
\centering
\begin{tabular}{@{}c@{\hspace{.5em}}c@{\hspace{.5em}}c@{}}
  \includegraphics[width=28mm,height=40mm]{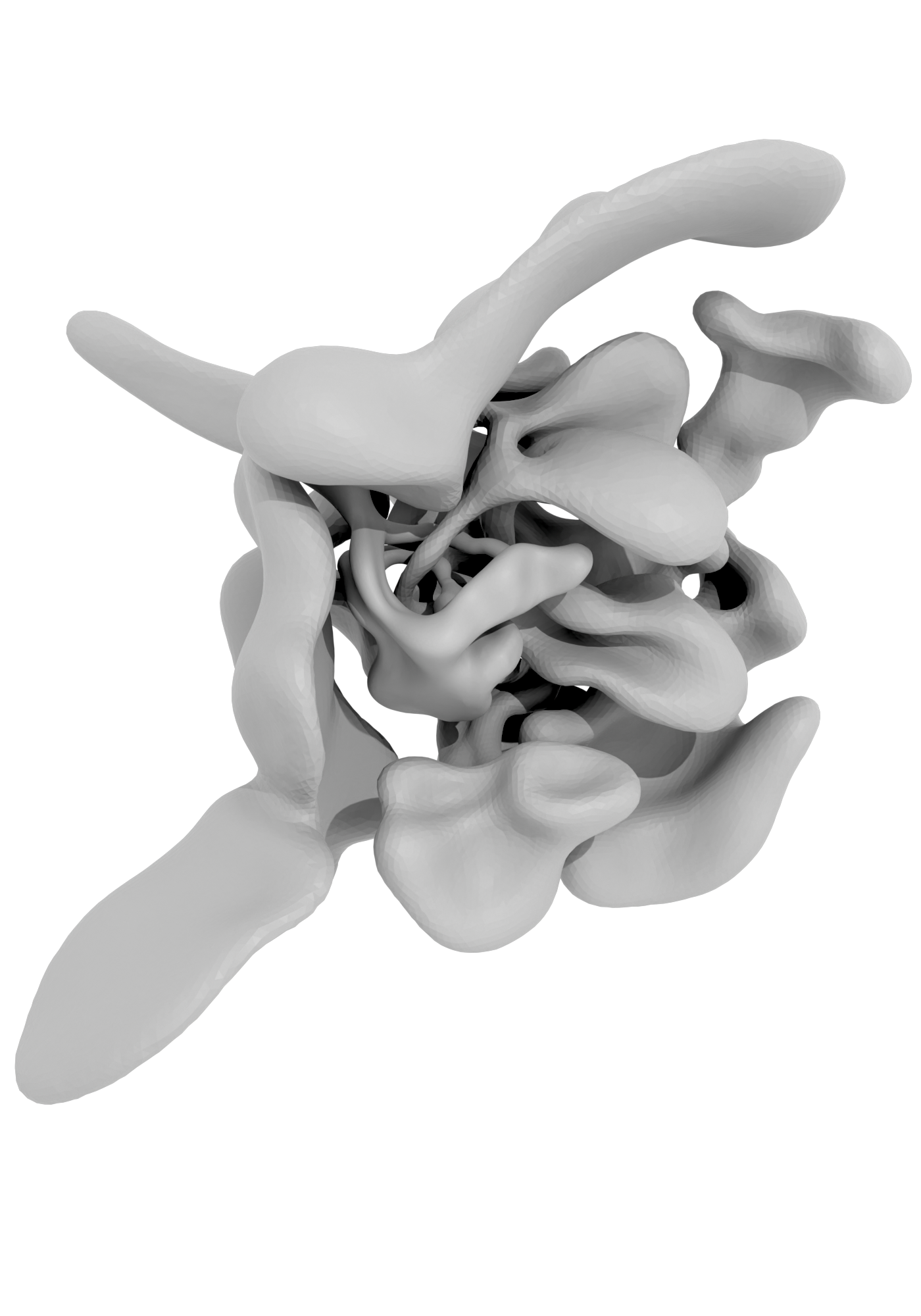} &
  \includegraphics[width=28mm,height=40mm]{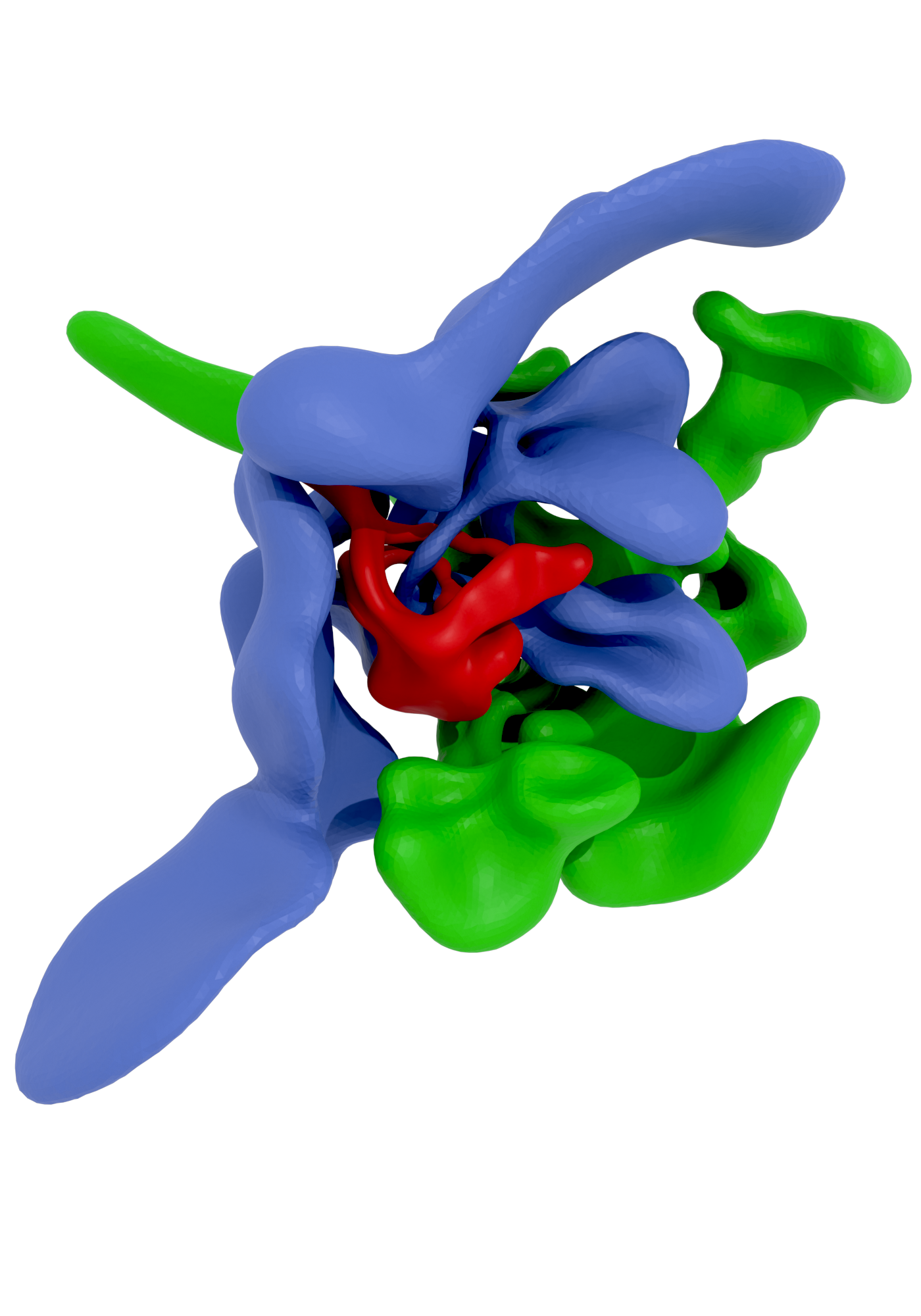} &
  \includegraphics[width=28mm,height=40mm]{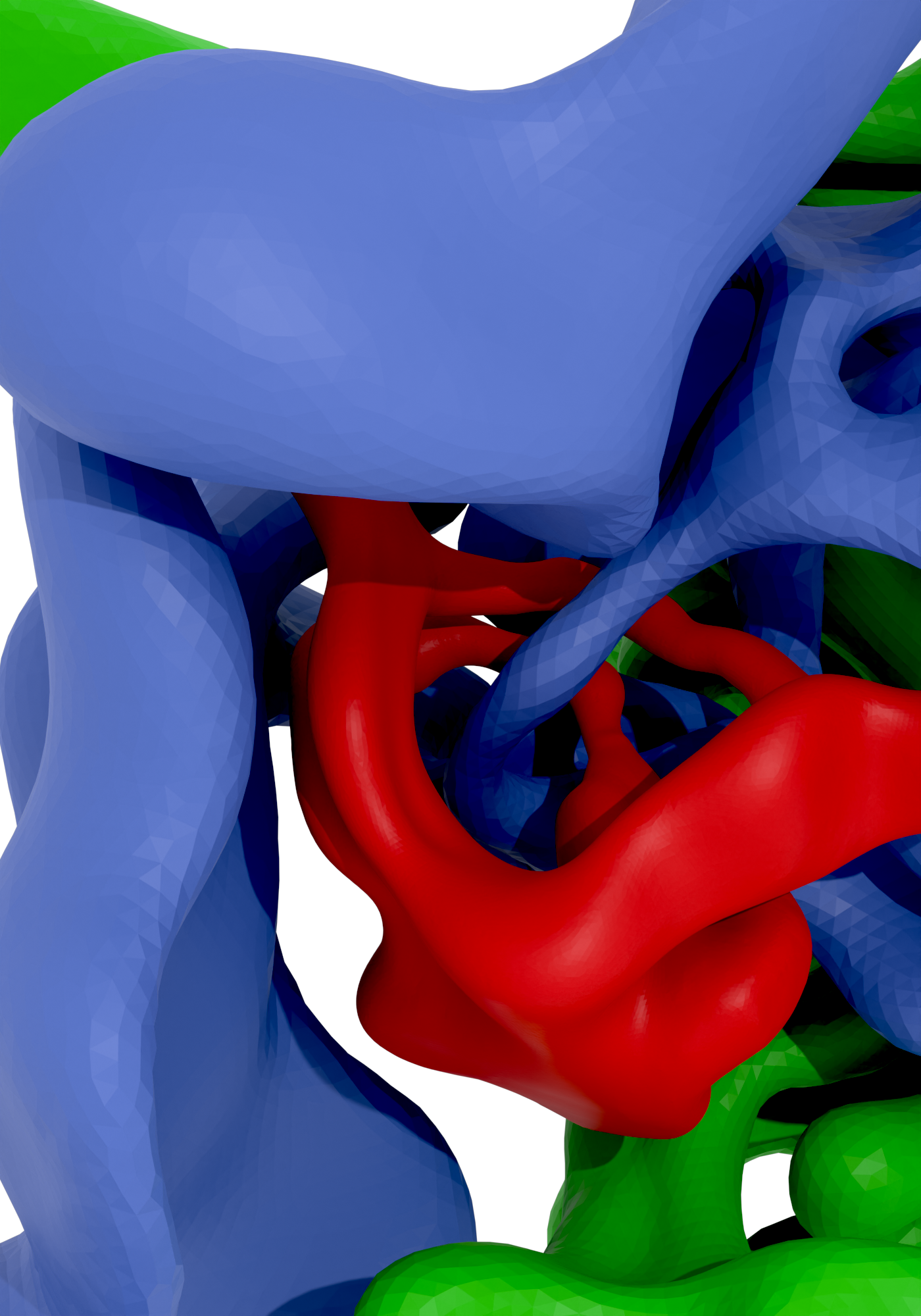} \\
  (a) & (b) & (c)
\end{tabular}
\caption{An example of a synthetic repulsive data sample comprising of 3 separate objects. Image (a) shows the generated output mesh. Image (b) shows the mesh coloured with genus 0 (green), genus 2 (red) and genus 3 (blue). Image (c) is a closeup of (b) to assist with link identification. Visualisation of data was performed in Blender 3.0.1. \cite{blender3.0.1}}\label{coloredmesh}
\end{figure}

Certain seed objects started linked which was preserved throughout the growth process (see Fig~\ref{coloredmesh}). This was an important dataset feature to add complexity as the labeling of objects should remain unaffected by these links.

The dataset consists of 5725 training samples, 1610 validation samples and 965 test samples. Each sample could contain 1-3 unique objects with each object having a genus of 0-3. To reduce training bias there is equal class representation for the amount of objects in the scene and equal quantities of each genus object. 

\begin{figure}[htbp]
\centering
\begin{tabular}{@{}c@{\hspace{.5em}}c@{}}
  \includegraphics[width=40mm,height=40mm, trim=0 50 0 50,clip]{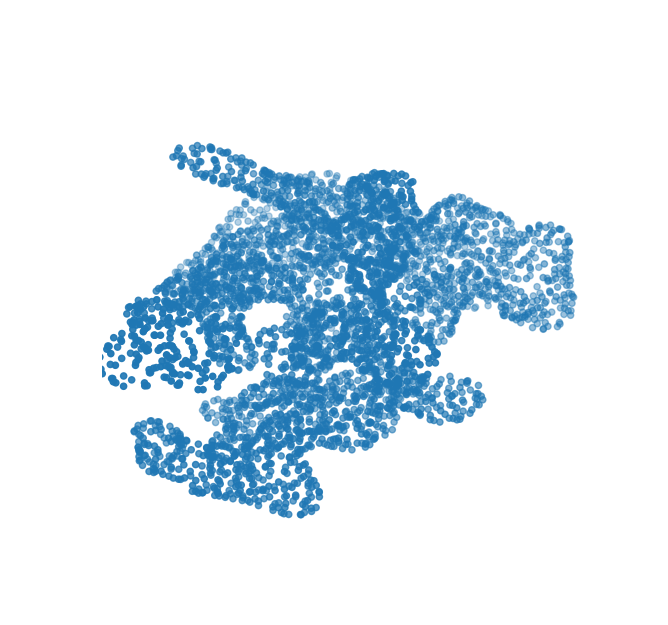} &
  \includegraphics[width=40mm,height=40mm, trim=0 50 0 50,clip]{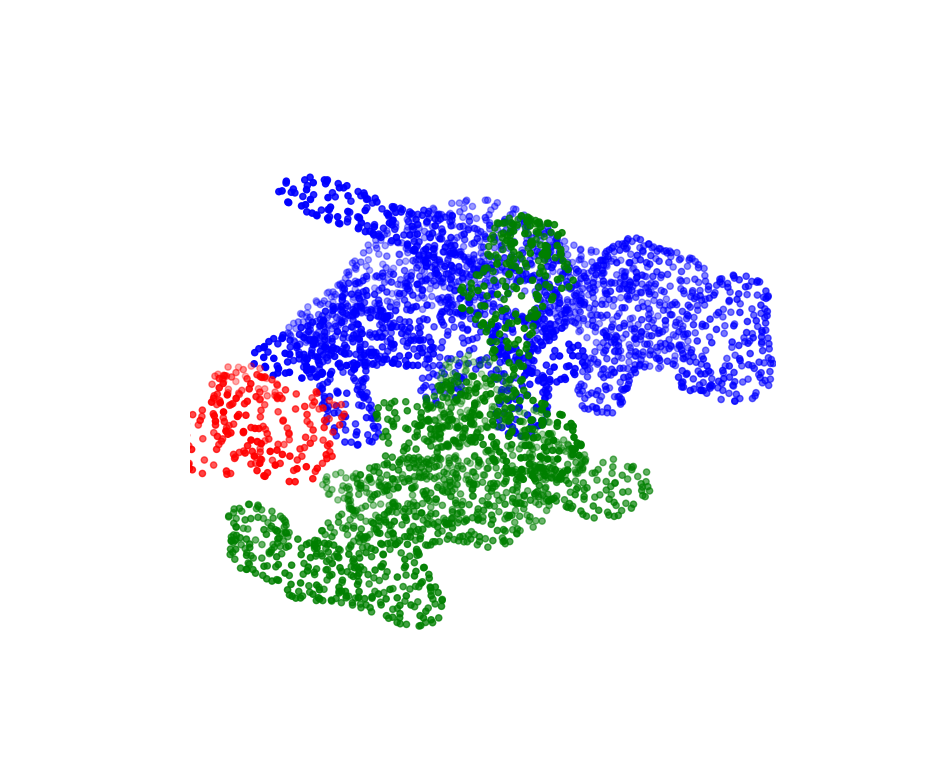} \\
  (a) & (b)
\end{tabular}
\caption{An example of a synthetic repulsive data sample comprising of 3 separate objects sampled at 4096 point count. Image (a) shows the point cloud input information and Image (b) visualises the target genus annotations via colour with genus 0 (red), genus 1 (green) and genus 2 (blue). }\label{dataset:pointcloud}
\end{figure}

While the dataset was generated as a mesh, this was less common for real-world application data and offered additional normal and boundary information that could provide an undesirable advantage. To counter this the training, validation and evaluation was performed on a point cloud variant. The generated meshes were sampled uniformly with a point count of 4096. Each point was labeled with the parent object genus for segmentation. An example sampled at 4096 with labels can be seen in Fig~\ref{dataset:pointcloud}. 

\subsection{Training} \label{training}
Each network was trained on the complete training dataset for 100 epochs. The validation dataset was evaluated at the end of each epoch for best model checkpointing. The test set was used for unbiased evaluation on the best validation model after training had finished.
The initial learning rate was 0.001 which was reduced by a factor of 10 every 25 epochs with a batch size of 64.

In order to enhance the size of the dataset and mitigate overfitting, various data augmentation techniques were applied to the training set. These techniques include:
\begin{itemize}
  \item $50\%$ chance to mirror each of the 3 axis.
  \item Full $2\pi$ rotation around each axis.
  \item Anisotropic scale deformation of each axis in the range $\left\{0.5, 1.5\right\}$, uniformly distributed.
  \item Shift of $\pm25$ in each axis independently, uniformly distributed.
  \item Gaussian noise jitter applied to each points position with a standard deviation of 0.025. 
\end{itemize}

These augmentations were applied globally to all of the points within a given scene to prevent class-altering deformations.
 
\section{Results and Discussion}
A common problem in machine learning is knowing whether networks are actually learning to extract desired features from a dataset or are instead learning to estimate the output based on other contextual clues or artifacts. 
With this in mind, it was important to create a dataset where unique, random homeomorphic deformations could occur to attempt to mitigate these artificial features. This could aid in evaluating neural networks' true ability to assess topology; particularly addressing the question of whether the labels were extrapolated via artifacts from a succession of similar objects existing in both the train and test sets.

Genus segmentation also poses some unique challenges in that objects with the same topological features can take infinite forms in Euclidean space. This prevents the network from learning global shape alone and requires a deep understanding of the relationship between points.

\subsection{Experiments}
Training and evaluation on the `Repulse' dataset was conducted with 4096 points per scene.
Three networks for semantic segmentation were used: PointNet++, Point Transformer and RepSurf-U.

\begin{table}[h!]
\centering
\caption{Summary metrics of neural network models}
\begin{tabular}{cccc}
\toprule
Model & mIoU & mAcc & OA \\
\midrule
Point Transformer & 68.80 & 81.18 & 80.99 \\
RepSurf-U & 64.35 & 77.77 & 78.03 \\
PointNet++ & 57.67 & 72.88 & 72.61 \\
\bottomrule
\end{tabular}
\label{results:summarymetrics}
\end{table}

\begin{table}[h]
\centering
\caption{Class accuracy evaluated on Repulse dataset}
\begin{tabular}{cccc}
\toprule
Model & Genus & IoU & Acc \\
\midrule
\multirow{4}{*}{Point Transformer}
 & 0 & 78.80 & 87.89 \\
 & 1 & 70.12 & 81.82 \\
 & 2 & 56.25 & 72.87 \\
 & 3 & 70.03 & 82.13 \\
\midrule
\multirow{4}{*}{RepSurf-U} 
 & 0 & 75.54 & 83.88 \\
 & 1 & 58.85 & 72.89 \\
 & 2 & 53.97 & 73.51 \\
 & 3 & 69.84 & 80.80 \\
\midrule
\multirow{4}{*}{PointNet++} 
 & 0 & 67.52 & 77.73 \\
 & 1 & 57.07 & 76.33 \\
 & 2 & 45.00 & 63.51 \\
 & 3 & 61.11 & 73.94 \\
\bottomrule
\end{tabular}
\label{results:classmetrics}
\end{table}

\subsection{Discussion}
The obtained results demonstrated mean IoU and mean accuracy comparable to those achieved on datasets such as S3DIS (see Tables \ref{introduction:networkacc:table} and \ref{results:summarymetrics}) which is designed for object segmentation in building interior spaces. Interestingly, the overall accuracy metric on S3DIS was higher than on the Repulse set. It is speculated that the complexity of the dataset and unique challenges posed by the nature of the features may contribute to this difference. A tighter grouping between mIoU, mAcc, and OA which was seen in the Repulse experiments can also indicate a closer accuracy grouping between classes.

An observation that the accuracies vastly exceeded random distribution when applied to data with homeomorphic deformation is supportive for neural network TDA. This suggests that these networks are capable of assessing the connectivity and structure of the samples rather than purely assessing the local/global geometric features like the surface shape of known object classes. 

Such concepts are challenging to evaluate from existing studies as there were often correlations between object classification and topological features, see Section~\ref{relatedworks}. For example samples of varying coffee cups appearing in both training and evaluation datasets. This can make it unclear whether the networks encode topological labels into different object classifications, adjust learned persistence diagrams or images in response to changes in object shape, or perform a desirable topological analysis. 

\begin{figure}[htbp]
\centering
\begin{tabular}{@{}c@{\hspace{.5em}}c@{}}
  \includegraphics[width=40mm,height=40mm, trim=0 50 0 50,clip]{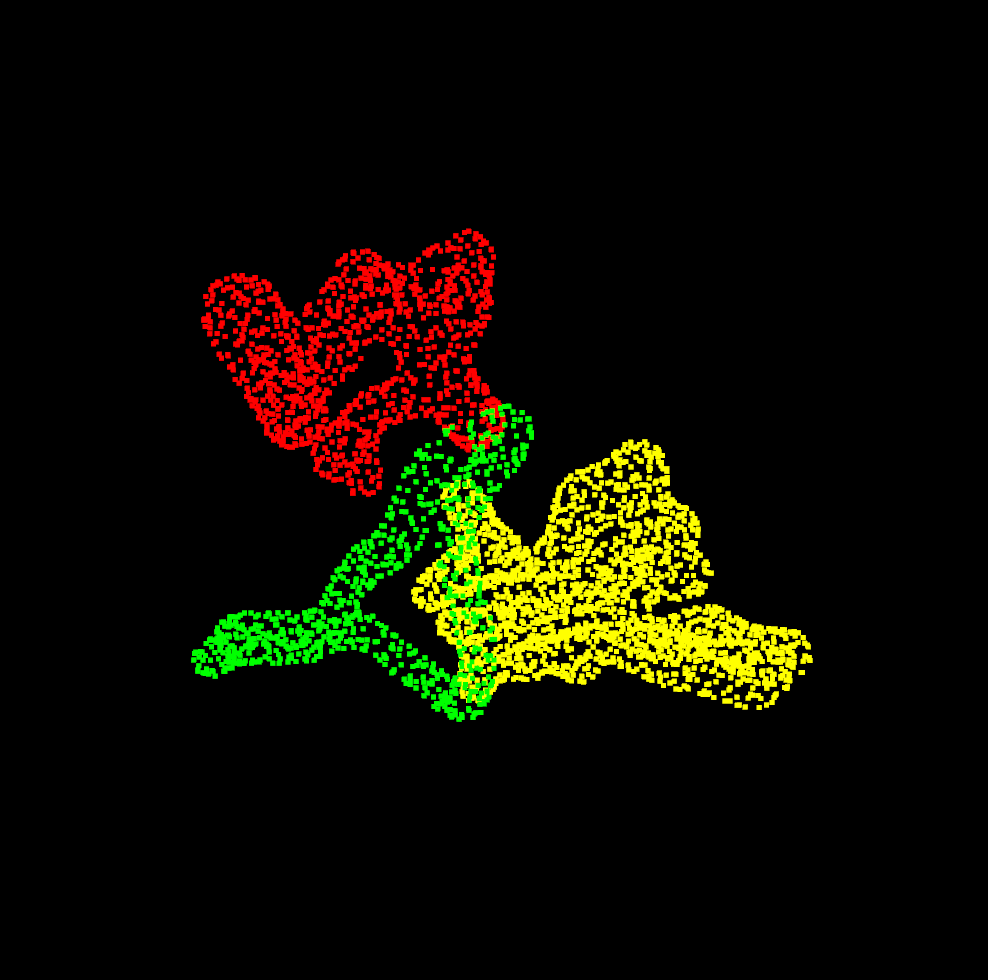} &
  \includegraphics[width=40mm,height=40mm, trim=0 50 0 50,clip]{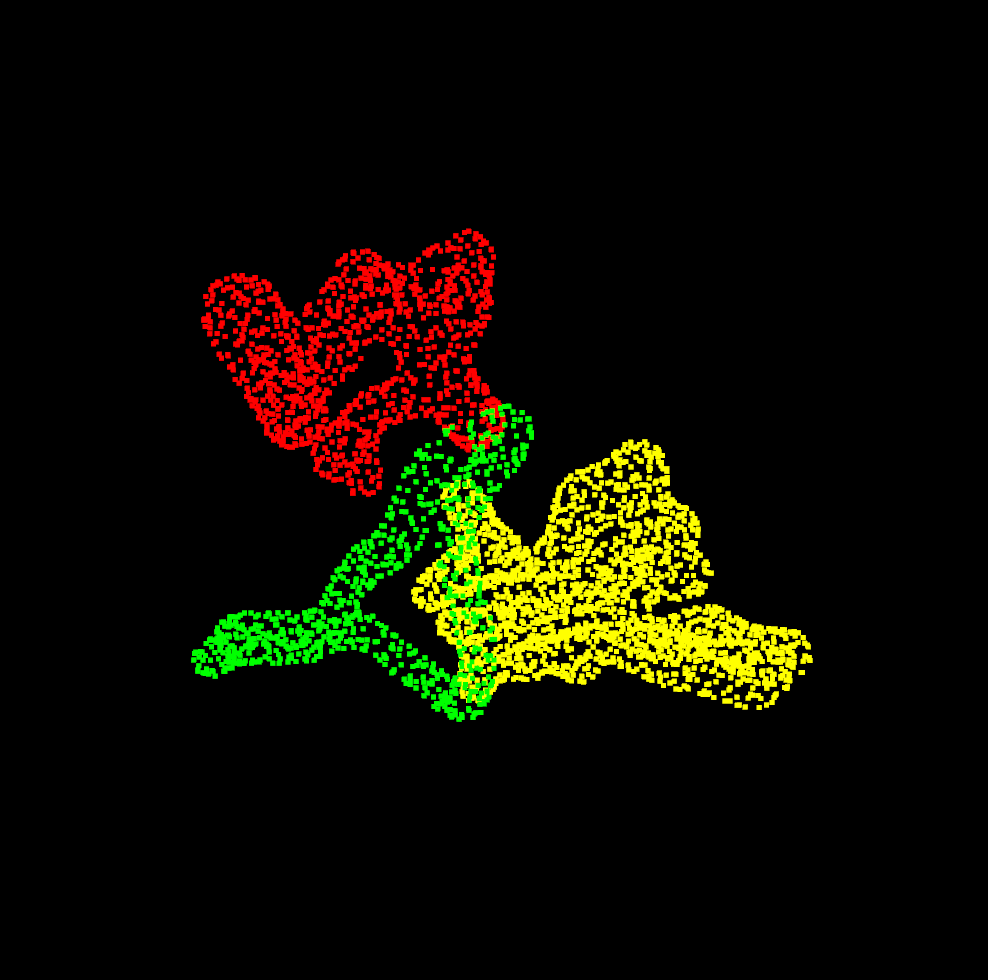} \\
\end{tabular}
\begin{tabular}{@{}c@{\hspace{.5em}}c@{}}
  \includegraphics[width=40mm,height=40mm, trim=0 50 0 50,clip]{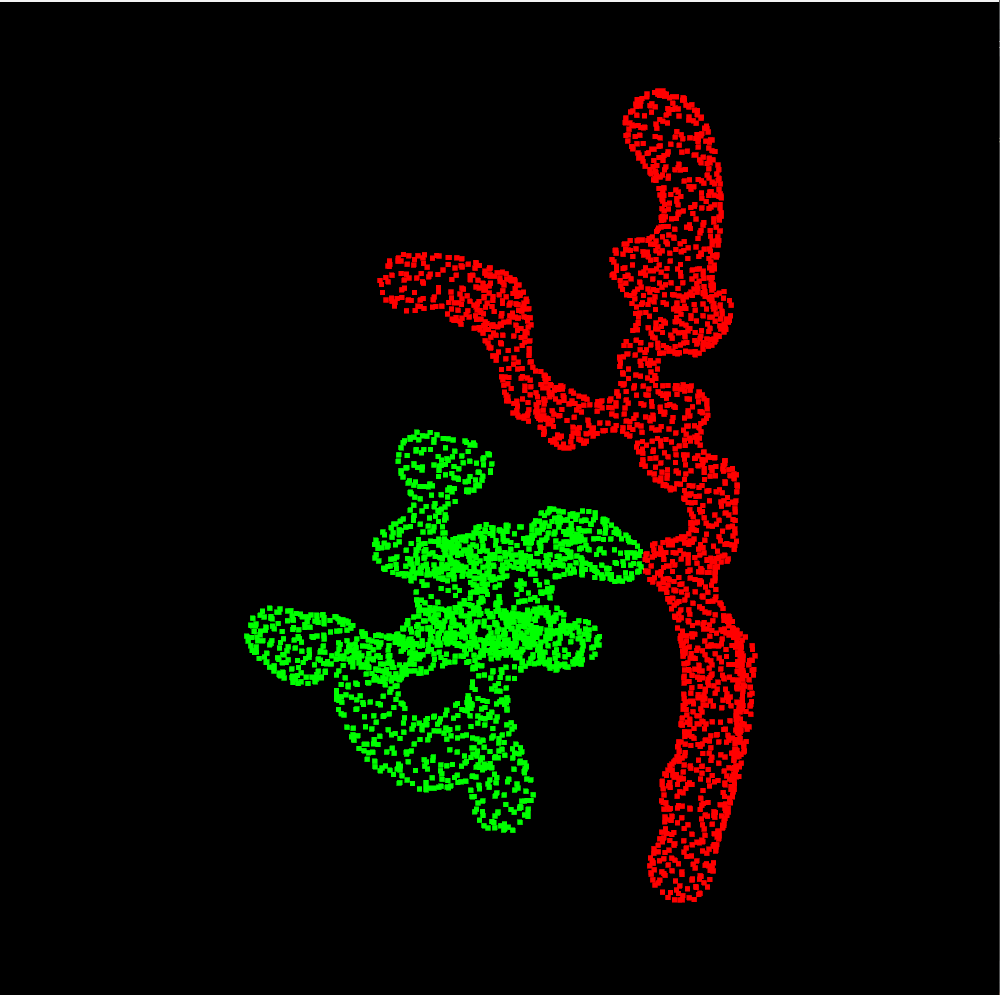} &
  \includegraphics[width=40mm,height=40mm, trim=0 50 0 50,clip]{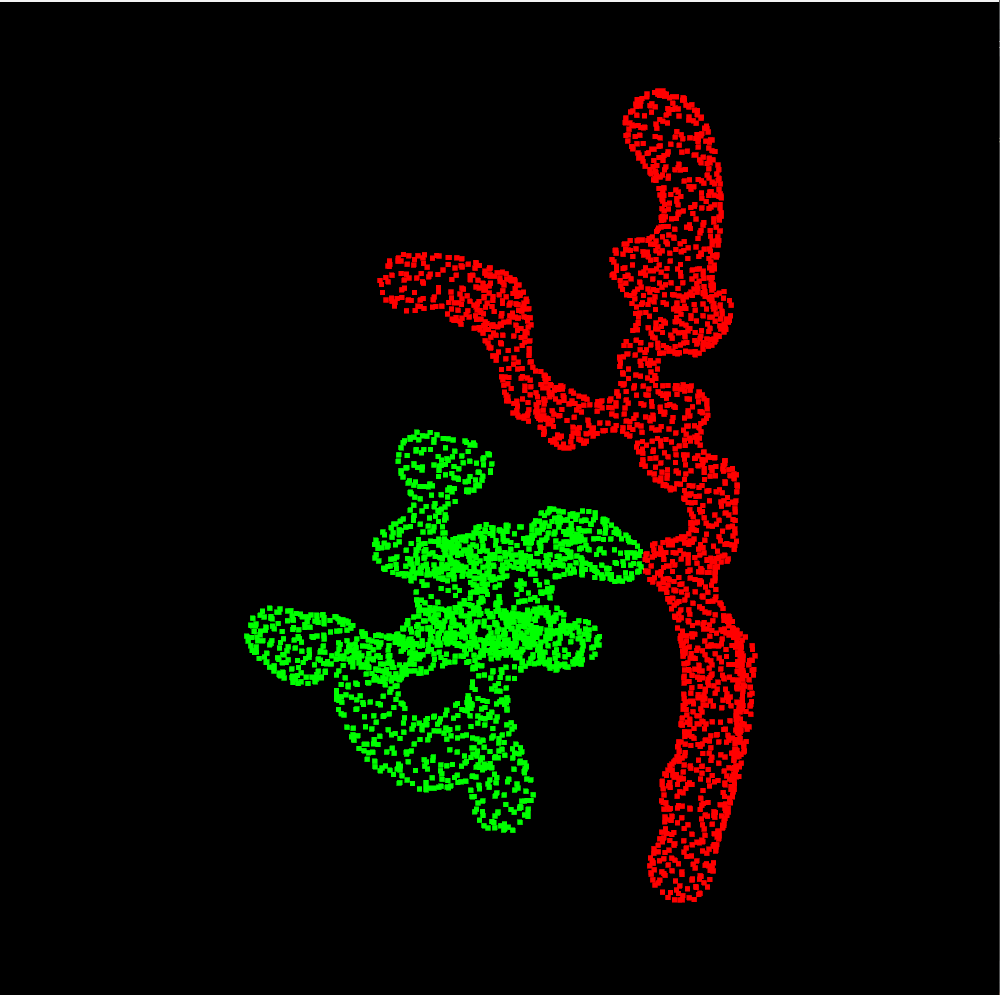} \\
\end{tabular}
\begin{tabular}{@{}c@{\hspace{.5em}}c@{}}
  \includegraphics[width=40mm,height=40mm, trim=0 50 0 50,clip]{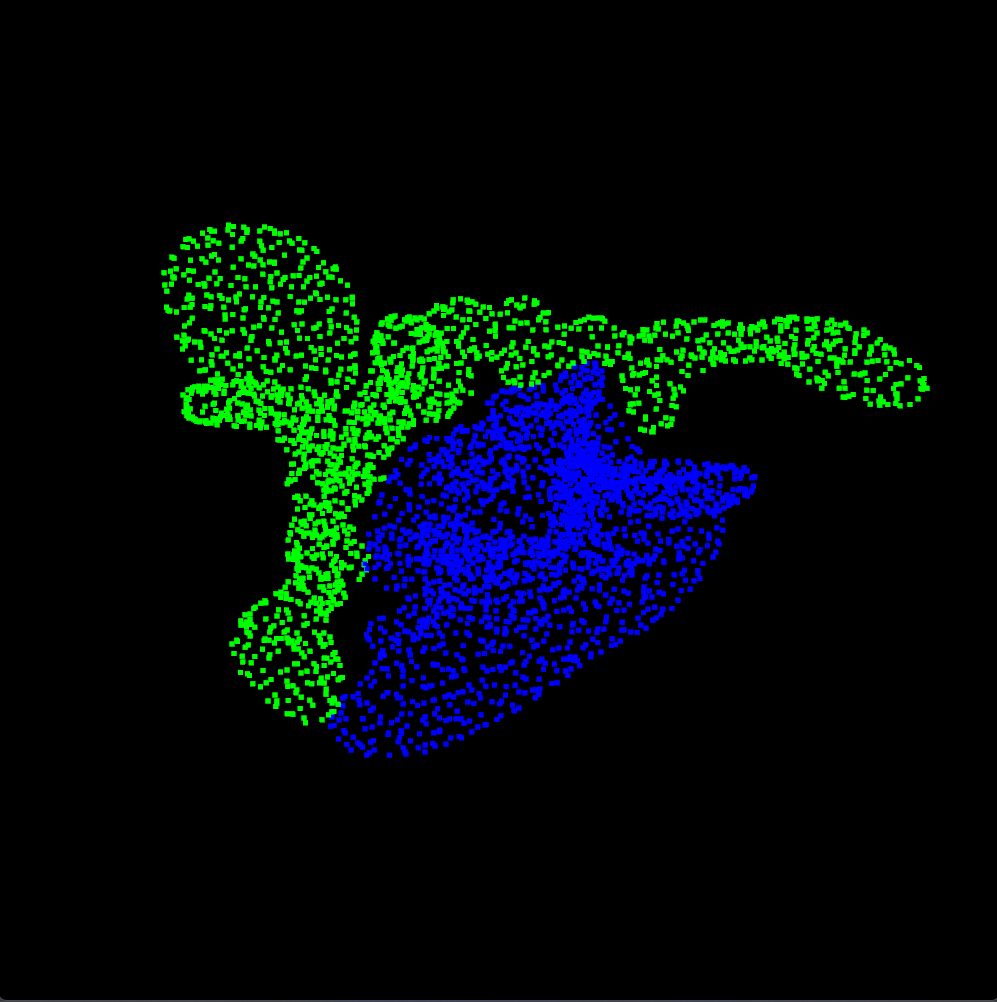} &
  \includegraphics[width=40mm,height=40mm, trim=0 50 0 50,clip]{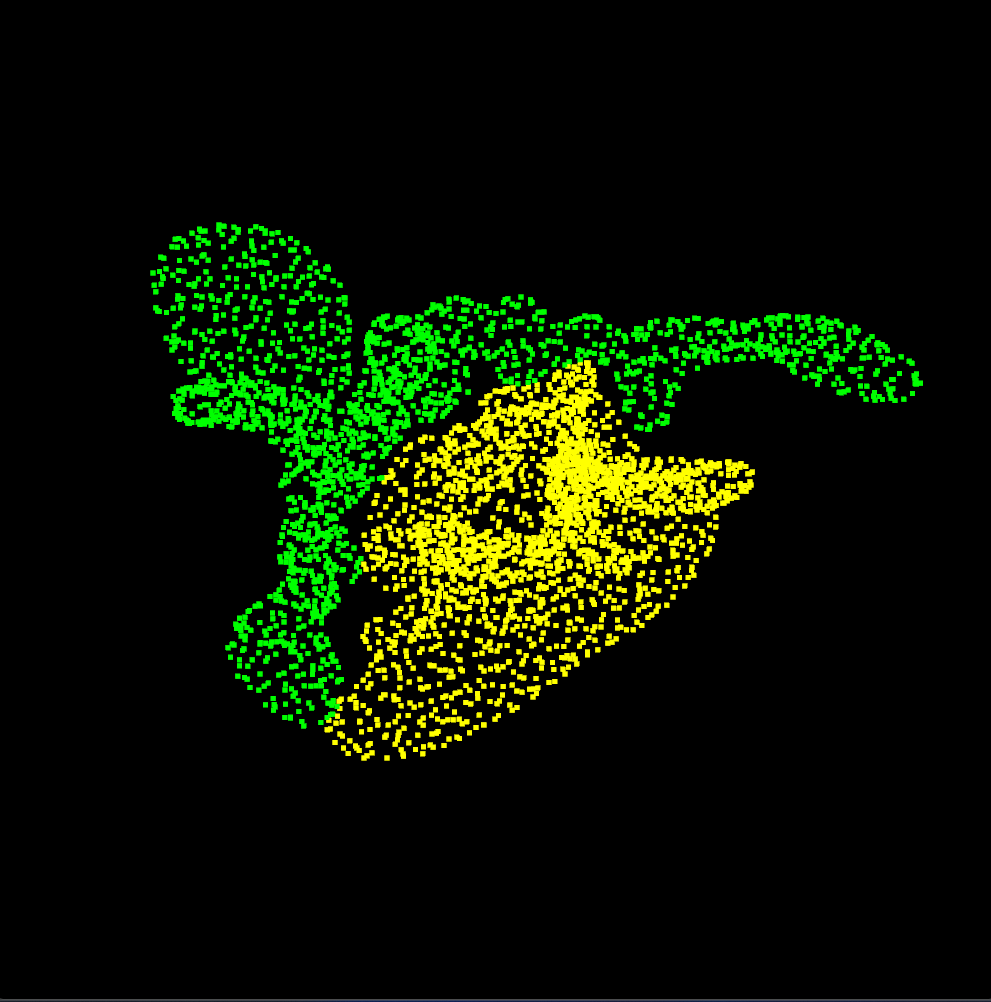} \\
\end{tabular}
\begin{tabular}{@{}c@{\hspace{.5em}}c@{}}
  \includegraphics[width=40mm,height=40mm, trim=0 50 0 50,clip]{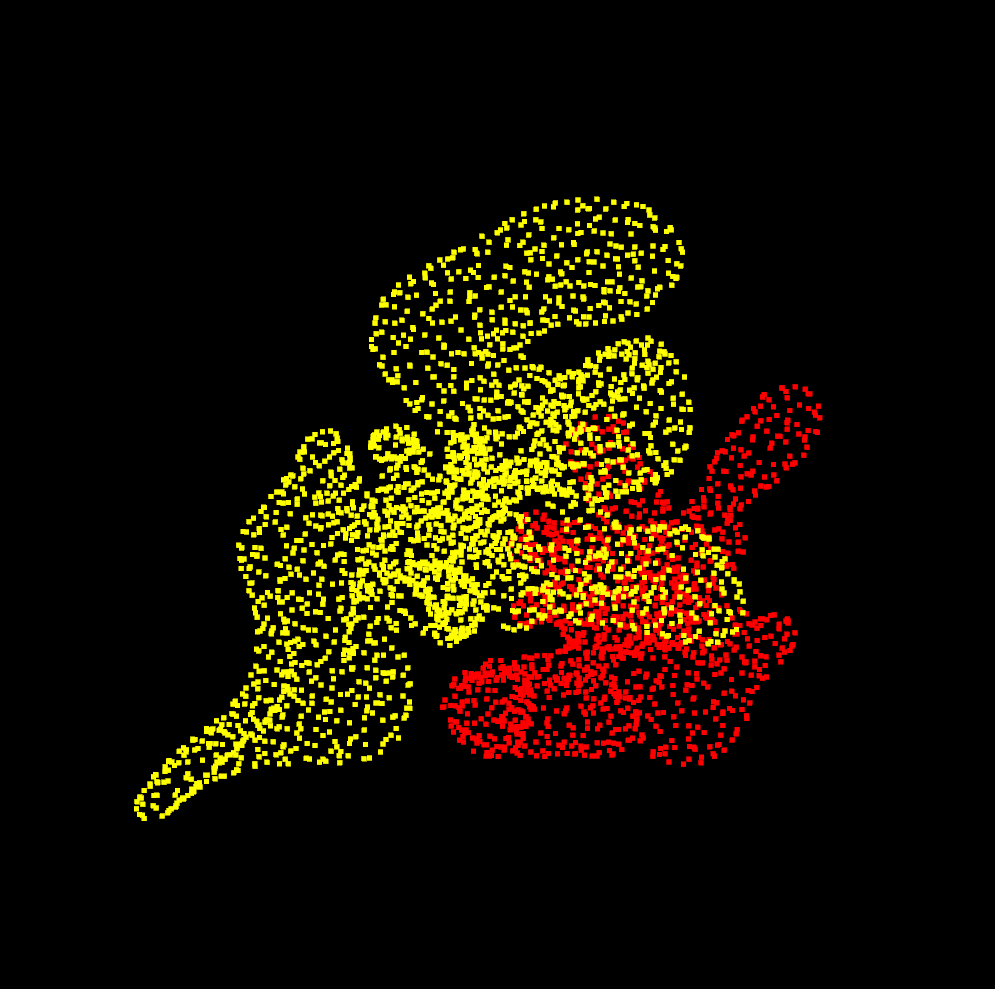} &
  \includegraphics[width=40mm,height=40mm, trim=0 50 0 50,clip]{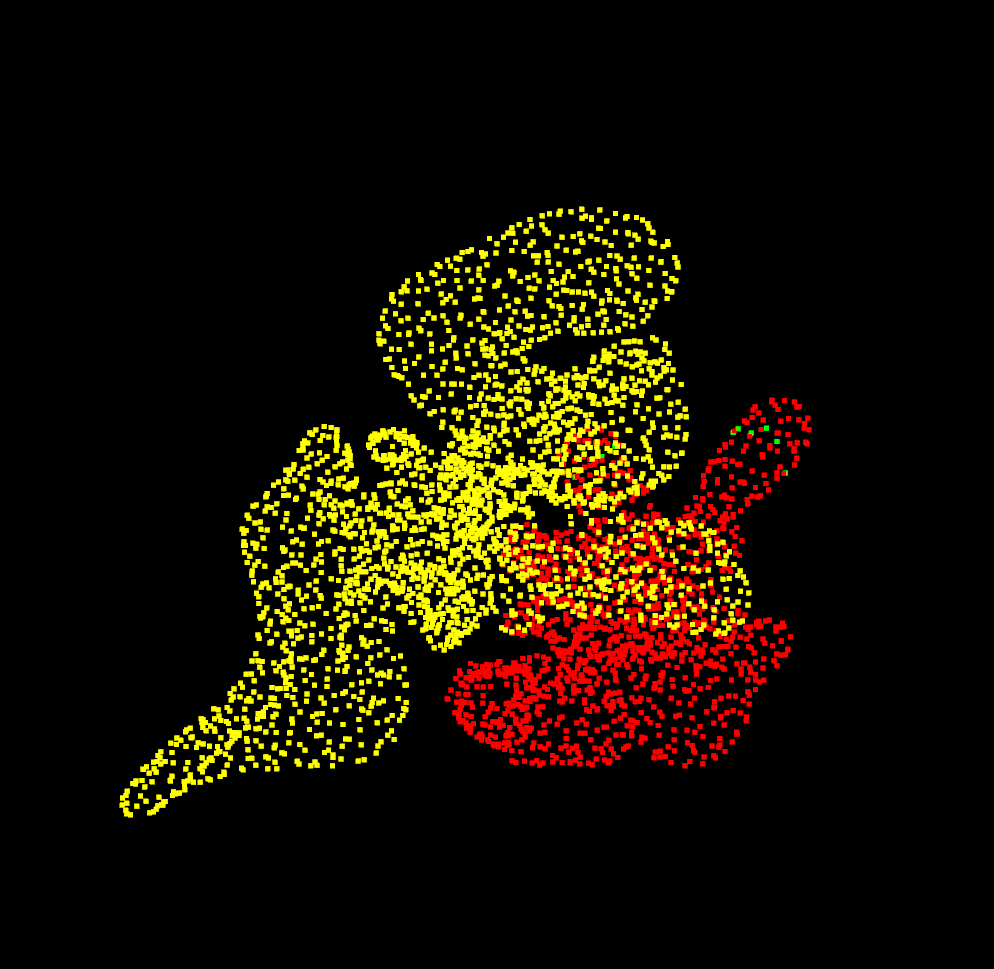}
\end{tabular}
\caption{Semantic segmentation output annotations for 4096 Repulse dataset. The left column shows the target annotations and the right column visualises the Point Transformer network output. Classes are coloured with genus 0 (red), genus~1 (green), genus 2 (blue) and genus 3 (yellow). }\label{results:outputs}
\end{figure}

Of the three networks trained Point Transformer achieved the highest accuracy which showed that transformer architectures with self-attention may be suitable for this topological segmentation task. Self-attention adjusts point significance with respect to a centroid point belonging to a local neighbour group. This may prioritise certain manifold properties such as curvature and connectivity, aiding in the topological analysis and increasing accuracy. Segmented output examples from this network are provided in Fig~\ref{results:outputs}. It is notable that the output in row 3 demonstrated a `clean' miss-classification in which the objects were successfully distinguished despite an incorrect hole count for the genus 2 object. A class IoU is determined by both the amount of points correctly labelled and the amount incorrectly labelled; $IoU(A, B) = \frac{|A \cap B|}{|A \cup B|}$, where $A$ and $B$ in this context represent correct and total classification sets. For this specific example, the lack of `bleeding' between close proximity objects offers an IoU of 100\% for genus 1, and 0\% for genus 2 resulting in a theoretical mIoU of 50\%.

RepSurf-U achieved similar accuracy however comprised of significantly less parameters than the Point Transformer network (0.976M vs 7.767M) which may be desirable for certain applications and hardware. 

The discrepancy in accuracy between PointNet++ and RepSurf-U seems to indicate that utilizing the umbrella curvature offered an advantage in topological data analysis as RepSurf-U utilizes PointNet++ as a backbone.

No explicit conclusion can be drawn from these preliminary experiments with respect to optimal architecture as there are many variants of MLPs and transformers as well as different training methods. Additionally, further adjustments to the hyper-parameters may show improvement to tune to this new topological task. It does however offer some insight into the construction of future TDA networks. 

\section{Conclusion}
Overall the pilot results of this study demonstrate that the concept of semantic segmentation for topological data analysis appears feasible. Further studies can be conducted into optimizing network architectures and data generation approaches to increase accuracy and applicability. 

The proposed approach is computationally efficient once the neural networks are trained and offers an advantage over persistent homology because it estimates the genus of each individual object in a scene and hence resolves ambiguities that are inherent to the Betti number evaluations of persistent homology as was pointed out in Section~\ref{ssub:background:toplogy}.

The new topological 3D image dataset was a crucial component of model training. It had a mix of genus and objects with unique shape and curvature which is a higher level of complexity and closer representation of real-world data than in previous studies. 

Limitations of this study include a finite object and genus count. The `Repulse' dataset had 1-3 objects with a genus of 0-3. In future studies this dataset could be greatly expanded to have larger signature variety. Additionally, this data could be further enhanced with additional visual variety by different environmental generation techniques and realistic post-processing that preserves the underlying structures.

Limitations also include a finite quantity of networks tested. Training more networks with greater variance in augmentation, training scheme, point counts and noise would allow further insight to be drawn regarding the effects of architecture and hyper-parameters on accuracy. 


Future research could also address detailed benchmarking and comparisons of persistent homology techniques and deep learning-based approaches on different image data and data formats. Both approaches display a range of advantages and disadvantages and the present study could only discuss some aspects of this emerging field of research.   

\bibliographystyle{IEEEtranS}
\bibliography{literature.bib}

\vspace{12pt}

\end{document}